\pdfoutput=1

\documentclass[11pt]{article}

\usepackage{acl}

\usepackage{subfigure}

\usepackage{times}
\usepackage{latexsym}
\usepackage{tabularx}
\usepackage{multirow}
\usepackage{booktabs} 
\usepackage{enumitem}
\usepackage{amsmath}

\usepackage{smile}
\usepackage{inconsolata}
\usepackage{graphicx}

\usepackage[utf8]{inputenc} 
\usepackage[T1]{fontenc}    
\usepackage{hyperref}       
\usepackage{url}            
\usepackage{booktabs}       
\usepackage{amsfonts}       
\usepackage{nicefrac}       
\usepackage{microtype}      
\usepackage{xcolor}         
\usepackage{multicol}
\usepackage{transparent}
\usepackage{array}
\usepackage{bm}

\usepackage{xspace}
\usepackage{cleveref}
\usepackage{colortbl}
\usepackage{listings}
\usepackage{bbding}

\definecolor{green}{HTML}{009B55}

\lstdefinestyle{mystyle}{
  basicstyle=\ttfamily,
  frame=single,
  breaklines=true,
  breakindent=0pt,
  backgroundcolor=\color{gray!10}, 
}
\crefname{section}{§}{§§}
\Crefname{section}{§}{§§}

\newcommand{\method}{\textsc{BMRetriever}\xspace}

\title{\method{}: Tuning Large Language Models as Better \\ Biomedical Text Retrievers}

\author{Ran Xu$^\heartsuit$\thanks{~Equal contribution.},  Wenqi Shi$^{\spadesuit}$\footnotemark[1], Yue Yu$^ \spadesuit$\footnotemark[1], Yuchen Zhuang$^{\spadesuit}$, Yanqiao Zhu$^{\diamondsuit}$ \\ \bf May D. Wang$^{\spadesuit}$, Joyce C. Ho$^{\heartsuit}$, Chao Zhang$^{\spadesuit}$, Carl Yang$^{\heartsuit}$ \\
\vspace{2pt}
$^{\heartsuit}$ Emory University~~~~$^{\spadesuit}$ Georgia Tech~~~~$^{\diamondsuit}$ UCLA \\
\texttt{\{ran.xu,joyce.c.ho,j.carlyang\}@emory.edu}, \texttt{yzhu@cs.ucla.edu}\\
\texttt{\{wqshi,yueyu,yczhuang,maywang,chaozhang\}@gatech.edu}
}

\begin{document}
\maketitle
\begin{abstract}
Developing effective biomedical retrieval models is important for excelling at knowledge-intensive biomedical tasks but still challenging due to the deficiency of sufficient publicly annotated biomedical data and computational resources.
We present \method, a series of dense retrievers for enhancing biomedical retrieval via unsupervised pre-training on large biomedical corpora, followed by instruction fine-tuning on a combination of labeled datasets and synthetic pairs.
Experiments on 5 biomedical tasks across 11 datasets verify \method's efficacy on various biomedical applications.
\method also exhibits strong parameter efficiency, with the 410M variant outperforming baselines up to 11.7 times larger, and the 2B variant matching the performance of models with over 5B parameters.
The training data and model checkpoints are released at \url{https://huggingface.co/BMRetriever} to ensure transparency, reproducibility, and  application to new domains.

\end{abstract}

\section{Introduction}
\label{sec:intro}

In the field of biomedicine, the ability to effectively retrieve knowledge from external corpora is crucial for large language models (LLMs) to excel at biomedical NLP tasks~\citep{lewis2020retrieval}. By tapping into up-to-date domain knowledge, retrieval-augmented LLMs have demonstrated promising results in various biomedical downstream applications, including knowledge discovery~\citep{frisoni-etal-2022-bioreader}, question answering~\citep{zhang2024raft,yu2024rankrag}, and clinical decision-making~\citep{naik-etal-2022-literature,shi2023retrieval,xu2024ram}.

\begin{figure}[t]
    \centering    \includegraphics[width=0.9\linewidth]{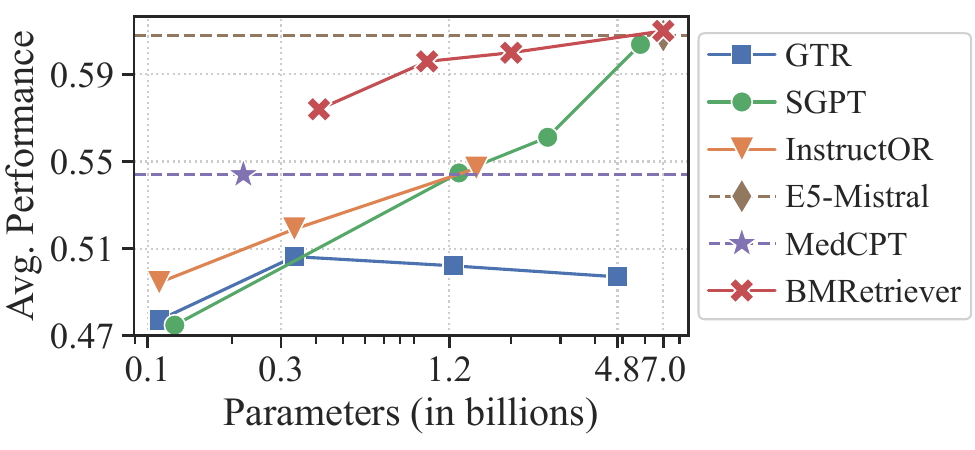}
    \vspace{-1.5ex}
    \caption{The average performance of \method on 5 popular biomedical search tasks compared to baselines with different parameters. X-axis in log scale.}
    \vspace{-2.8ex}
    \label{fig:perf}
\end{figure}

Several works have designed specialized retrieval models for biomedical domains
~\cite{mohan-etal-2017-deep,liu-etal-2021-self,jin2023medcpt,luo2022improving,singh-etal-2023-scirepeval,zhang-etal-2023-pre}. However, these models are typically built upon BERT-series models, which have limited representative power. 
Besides, they often rely on proprietary data (e.g., private search logs or patient records), making it challenging to scale them up to accommodate larger models effectively due to privacy concerns. 
While recent studies in the general domain have improved neural retrieval models via scaling up model size~\cite{ni-etal-2022-large,wang2023improving} and training data~\citep{izacard2022unsupervised,e5,lin-etal-2023-train}, 
adapting such models to the biomedical domain may lead to suboptimal performance due to the distribution shift issue~\citep{beir}. 
Developing large-scale retrieval models dedicated to the biomedical domain without requiring massive proprietary datasets remains crucial yet challenging.

In this work, we propose \method, a series of dense text retrievers at various scales using LLMs as backbones to improve biomedical retrieval performance.
Firstly, we inject biomedical knowledge into \method by unsupervised contrastive pre-training on a \emph{large-scale} unlabeled biomedical corpora, which comprises an extensive and diverse collection of data, with rich biomedical background knowledge invaluable for domain-specific understanding~\citep{lala2023paperqa,xiong2024benchmarking}.
Besides, unlabeled corpora are readily accessible, overcoming the bottleneck of scarce annotated data that often plagues specialized domains.
Pre-training on them allows us to adapt our models to the biomedical domain, equipping them with necessary linguistic patterns and terminology.

To further boost the embedding quality and align the retriever with downstream applications, we conduct instruction fine-tuning with \emph{high-quality} labeled datasets. 
Specifically, we gather various \textit{public human-annotated} biomedical retrieval tasks, such as medical question-answering (QA) and dialogue pairs, and create instructions for each to improve \method with task-specific understanding.
Given the relatively small sample size and limited task types in public biomedical datasets, we further leverage the powerful GPT models to generate additional synthetic retrieval tasks under various scenarios with query and passage pairs to augment training samples and diversify instructions.
This allows the model to acquire a comprehensive understanding of biomedical retrieval tasks and facilitates its generalization across various downstream tasks and input formats.

We conduct extensive experiments across \emph{five} tasks on \emph{eleven} biomedical datasets to demonstrate the strong performance of \method.
As shown in Figure~\ref{fig:perf}, \method outperforms existing dense retrievers with orders of magnitude more parameters: with 410M parameters, it surpasses the performance of GTR-4.8B~\citep{ni-etal-2022-large} and SGPT-2.7B~\citep{muennighoff2022sgpt}, which have $7\times$ more parameters. 
At the 7B scale, \method outperforms the recently proposed E5-Mistral~\citep{wang2023improving}, which uses extra-large batch-size and nonpublic data mixture.
In addition, \method presents a lightweight yet high-performing domain adaptation solution, with its 1B variant achieving more than 98\% performance of E5-Mistral using only 14.3\% of parameters. 
Our contribution can be summarized as follows:
\vspace{-1.3ex}
\begin{itemize}[leftmargin=0.3cm]
\setlength\itemsep{-0.2em}
\item We develop a family of \method models ranging from 410M to 7B parameters, achieving efficient scaling via a two-stage framework to improve biomedical text retrieval performance.
\item We assess \method's efficacy with an extensive evaluation against 18 baselines on 5 tasks across 11 biomedical datasets. Results demonstrate \method's parameter efficiency yet strong domain adaptation capabilities, achievable within academic computational budgets.
\item \method ensures transparency, reproducibility, and potential generalization to additional domain-specific adaptations by providing a detailed training recipe with public datasets and accessible model checkpoints.
\end{itemize}

\section{Related Work}
\label{sec:related}
Earlier research explores various approaches for learning representations suitable for text retrieval~\citep{deerwester1990indexing, huang2013learning}. 
More recently, several studies introduce dual-encoder architectures based on BERT for dense retrieval~\cite{dpr,ance,gao-callan-2022-unsupervised,izacard2022unsupervised}. 
With the advent of LLMs with billions of parameters, several studies attempt to scale up model size~\citep{ni-etal-2022-large,neelakantan2022text}, often fine-tuned on multi-task instruction data~\citep{asai-etal-2023-task,su-etal-2023-one,wang2023improving}. However, the benefit of scaling up is more pronounced for general domain datasets where massive annotated data are available.

To design effective retrievers for specialized domains, several works propose continuously pre-train the retrieval model on domain-specific corpora~\citep{yu-etal-2022-coco,zhang-etal-2023-pre} or fine-tuning the model on proprietary search datasets~\citep{mohan-etal-2017-deep,jin2023medcpt}. 
On the other hand, synthetic data has also been used to improve the generalization ability of dense retrieval model~\citep{wang-etal-2022-gpl,jiang-etal-2023-noisy,wang2023improving,zhang2024arl2}. 
Despite these advancements, how to combine public, open data to formulate a dataset curation recipe for adapting LLMs as high-performing biomedical retrievers remains unresolved.
Our method efficiently integrates diverse supervision signals for biomedical retrieval model training, which achieves better performance than baselines trained with more data.

\section{Method}
\label{sec:method}
 \vspace{-0.5ex}
\begin{figure*}[t]
    \centering    
    \includegraphics[width=0.99\linewidth]{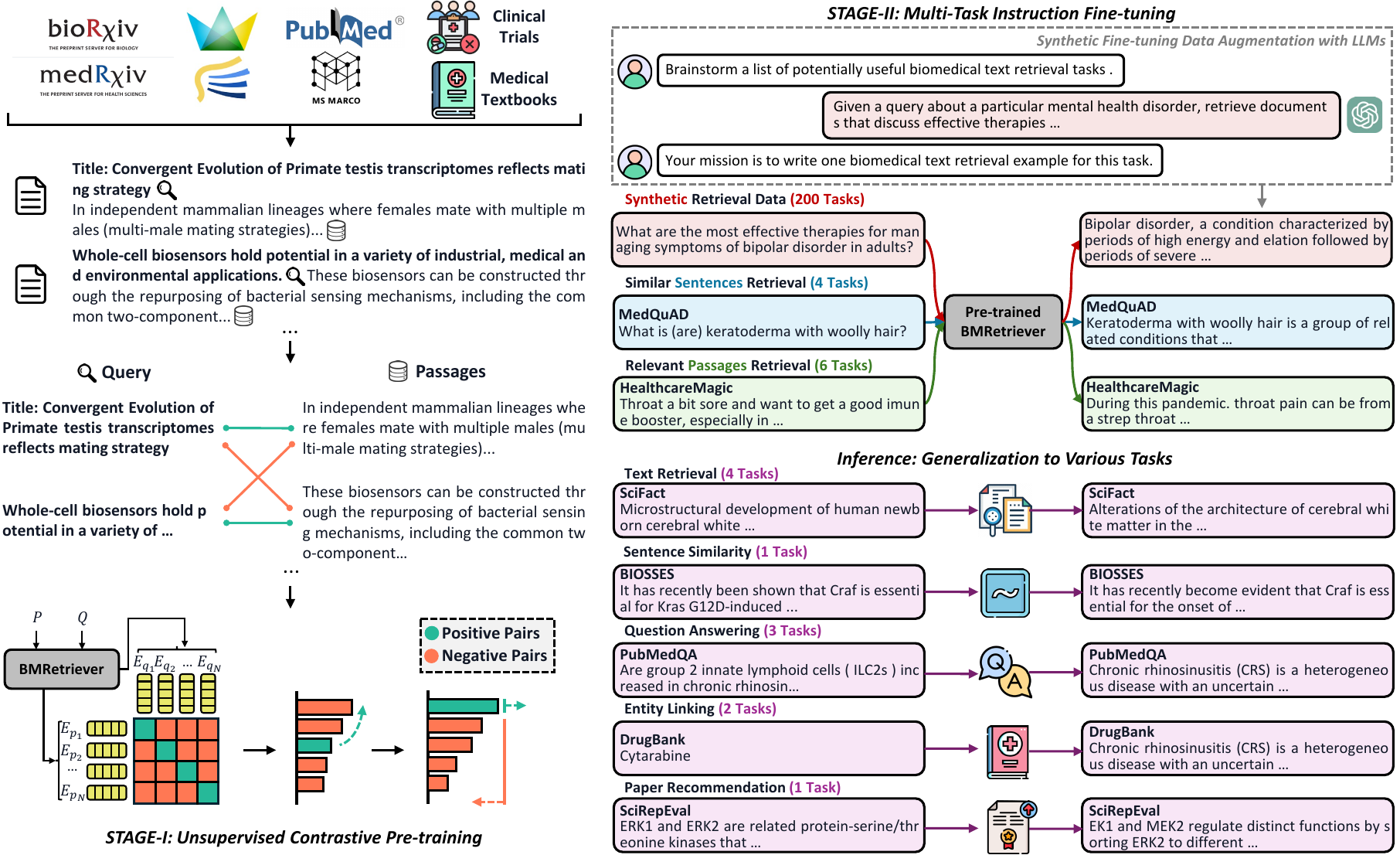}
    \caption{The overview of the two-stage pre-training framework in \method. Stage I performs unsupervised contrastive pre-training on large-scale biomedical query-passage pairs, while Stage II conducts instruction fine-tuning using diverse labeled data, including synthetic examples generated by LLMs, to adapt \method to various biomedical downstream tasks.\vspace{-0.5ex}}
    \vspace{-0.8ex}
    \label{fig:overview}
\end{figure*}

\begin{table}[t!]
\centering
\resizebox{\linewidth}{!}{
    \begin{tabular}{lcccc}
       \toprule
       \bf Parameters & 410M & 1B & 2B & 7B \\
       \midrule 
        \bf Backbone & Pythia~\shortcite{pythia} & Pythia~\shortcite{pythia} & Gemma~\shortcite{team2024gemma} & BioMistral~\shortcite{labrak2024biomistral}\\
        \bf Model Layers & 24 & 16 & 18 & 32 \\
        \bf Embedding Dim. & 1024& 2048 & 2048 & 4096 \\
       \bottomrule
    \end{tabular}
}
	\caption{An overview of \method. \vspace{-1.5ex}}
 \vspace{-1ex}
	\label{tab:method_overview}
\end{table}

\method{} leverages the pre-trained autoregressive transformer as the backbone, taking advantage of the availability of various model sizes within this model family. This flexibility allows us to scale up the retrieval model. Specifically, we utilize the publicly available autoregressive transformers with 410M, 1B, 2B, and 7B parameters~\citep{pythia,team2024gemma,labrak2024biomistral}. 
Our model details are illustrated in Table~\ref{tab:method_overview}.

\subsection{Background of Dense Text Retrieval}
In dense retrieval~\citep{lee-etal-2019-latent,dpr}, the language model $\mathbf{E}$ is used to represent queries and passages in dense embeddings.
Denote the query $q$ and passage $p$ with the corresponding task instruction $I_q$ and $I_p$\footnote{The instruction format is in Appendix~\ref{app:dataset}.}, 
the embedding is calculated as $\be_q=\mathbf{E}\left(I_q \oplus q\right)$, $\be_p=\mathbf{E}\left(I_p \oplus p\right)$. The relevance score $\operatorname{sim}(q, p)$ is calculated with the dot product between query and passage embeddings:
\begin{equation}
\setlength{\abovedisplayskip}{6pt}
\setlength{\belowdisplayskip}{6pt}
\operatorname{sim}(q, p)=\be_q^{\top}\be_p.
\end{equation}
In this work, where autoregressive LLMs are used for $\mathbf{E}$, an \texttt{<EOS>} token is appended to the end of the query and passage. The embedding of the \texttt{<EOS>} token from the final layer of LLM is used as the representation for both queries and passages.

To effectively adapt \method to the biomedical domain,
a two-stage training procedure is proposed (see Figure~\ref{fig:overview}): (1) an unsupervised contrastive pre-training stage (\cref{sec:method-cpt}) using \emph{silver} query-passage pairs from extensive biomedical corpora, and (2) a fine-tuning stage (\cref{sec:method-ift}) using \emph{gold} labeled data from various tasks. 
The details of the two stages will be introduced in the following sections.

\subsection{Unsupervised Contrastive Pre-training}
\label{sec:method-cpt}

\noindent\textbf{Pre-training Corpus Collection.}
To provide \method with an initial understanding of biomedical contexts, we collect a diverse range of publicly available biomedical corpora, including \emph{biomedical publications}~\citep{chen2021litcovid, xiong2024benchmarking, lo-etal-2020-s2orc}, \emph{medical textbooks}~\citep{textbook}, as well as \emph{general-domain web corpus}~\citep{msmarco}, as detailed in Table~\ref{tab:pt_data}.

\noindent\textbf{Contrastive Pre-training.} 
We construct positive and negative query-passage pairs from raw unlabeled corpora to facilitate contrastive pre-training of the retrieval model. For positive pairs, we employ two strategies: (1) for corpora with titles, we treat the title as the query and the corresponding abstract as the passage; (2) for untitled corpora, we randomly sample two disjoint passages from documents, using one as the query and the other as the passage~\citep{izacard2022unsupervised}. To obtain negative pairs, we sample in-batch negatives~\citep{gillick-etal-2019-learning} where the passages from other pairs in the same batch serve as negative examples. 
With the collected pairs, we employ contrastive learning to distinguish the relevant query-passage pairs from the irrelevant ones.
For each mini-batch $\cB$, we leverage the InfoNCE loss as the pre-training objective to rank the positive text pairs $\{(q_i,p_i)\}_{i=1}^n$ higher than in-batch negative passages $\{p_{ij}^-\}_{j=1}^N$:
\begin{equation}
\setlength{\abovedisplayskip}{5.2pt}
\setlength{\belowdisplayskip}{5.2pt}
\cL_{\text{cpt}}=-\log \frac{{e}^{\operatorname{sim}\left({q_i}, p_i\right)/\tau}}{\sum_{j\in \cB} {e}^{\operatorname{sim}\left(q_i, p_j\right)/\tau}}.
\label{eq:cl}
\end{equation}
Contrastive pre-training improves the quality of representations by better aligning similar text sequences while ensuring the uniformity of unrelated text
sequences, which helps adapt the retrieval model to biomedical domains~\citep{gururangan-etal-2020-dont,yu-etal-2022-coco}.

\subsection{Supervised Instruction Fine-tuning}\label{sec:method-ift}
To further enhance the model's specialized domain knowledge and align the model with downstream application tasks, we conduct instruction fine-tuning, which integrates a diverse collection of retrieval tasks into the instruction tuning blend. 
We present a detailed procedure below.

\begin{table*}[t]
\centering
\renewcommand\arraystretch{0.96}
\resizebox{\linewidth}{!}{
    \begin{tabular}{l|ccc|ccccc|cc}
       \toprule
         \bf Task & \bf \multirow{2.5}{*}{Scale} & \bf \multirow{2.5}{*}{\# PT Pairs} & \bf \multirow{2.5}{*}{\# FT Pairs} & \multicolumn{4}{c|}{\bf Standard IR} & \bf Sent. Sim. & \bf \multirow{2.5}{*}{Avg. Retr.} & \bf \multirow{2.5}{*}{Avg. All}\\
         \cmidrule{1-1} \cmidrule{5-9}
         \bf Model & & & & \bf NFCorpus & \bf {SciFact} & \bf {SciDocs} & \bf {Trec-COVID} & \bf {BIOSSES} \\
       \midrule 
       \multicolumn{8}{l}{\bf Sparse Retrieval} \\ 
       \midrule
       BM25~\citep{bm25} & --- & --- & --- & 0.325 & 0.665 & 0.158 & 0.656 & --- & 0.451 & --- \\
       \midrule 
       \multicolumn{8}{l}{\textbf{Base Size} (< 1B)}\\
       \midrule 
       Contriever~\citep{izacard2022unsupervised} & 110M & 1B & 500K & 0.328 & 0.677 & 0.165 & 0.596 & 0.833 & 0.442 & 0.520 \\
       Dragon~\citep{lin-etal-2023-train} & 110M & --- & 28.5M & 0.339 & 0.679 & 0.159 & 0.759 & 0.819 & 0.484 & 0.551 \\
       \rowcolor{green!15} SPECTER 2.0~\citep{singh-etal-2023-scirepeval} & 110M & 3.3M & --- & 0.228 & 0.671 & --- & 0.584 & --- & --- & --- \\
       \rowcolor{green!15} SciMult~\citep{zhang-etal-2023-pre} & 110M & 5.5M & --- & 0.308 & 0.707 & --- & 0.712 & --- & --- & --- \\
       COCO-DR~\citep{yu-etal-2022-coco} & 110M & 15M & 500K & 0.355 & 0.709 & 0.160 & 0.789 & 0.829 & 0.503 & 0.567 \\
       SGPT-125M~\citep{muennighoff2022sgpt} & 125M & unknown & 500K & 0.228 & 0.569 & 0.122 &  0.703 & 0.752 & 0.406 & 0.475 \\
       \rowcolor{green!15} MedCPT~\citep{jin2023medcpt} & 220M & --- & 255M & 0.340 & 0.724 & 0.123 & 0.697 & 0.837 & 0.471 & 0.544 \\
       GTR-L~\citep{ni-etal-2022-large} & 335M & 2B & 662K & 0.329 & 0.639 & 0.158 & 0.557 & 0.849 & 0.421 & 0.506 \\
       InstructOR-L~\citep{su-etal-2023-one} & 335M & --- & 1.24M & 0.341 & 0.643 & 0.186 & 0.581 & 0.844 & 0.438 & 0.519 \\
       E5-Large-v2$^\dagger$~\citep{e5} & 335M & 270M & 1M & 0.371 & 0.726 & 0.201 & 0.665 & 0.836 & 0.491 & 0.560 \\
       BGE-Large$^{*\ddagger}$~\citep{chen2024bge} & 335M & 1.2B & 1.62M & 0.345 & 0.723 & 0.222 & 0.753 & 0.804 & \textbf{0.511} & \underline{0.569} \\
       \rowcolor{violet!10} \method-410M & 410M & 10M & 1.4M & 0.321 & 0.711 & 0.167 & 0.831 & 0.840 & \underline{0.508} & \textbf{0.574} \\
       \midrule 
       \multicolumn{8}{l}{\textbf{Large Size} (1B - 5B)}\\
      \midrule
       InstructOR-XL~\citep{su-etal-2023-one} & 1.5B & --- & 1.24M & 0.360 & 0.646 & 0.174 & 0.713 & 0.842 & 0.473 & 0.547 \\
       GTR-XL~\citep{ni-etal-2022-large} & 1.2B & 2B & 662K & 0.343 & 0.635 & 0.159 & 0.584 & 0.789 & 0.430 & 0.502 \\
       GTR-XXL~\citep{ni-etal-2022-large} & 4.8B & 2B & 662K & 0.342 & 0.662 & 0.161 & 0.501 & 0.819 & 0.417 & 0.497 \\
       SGPT-1.3B~\citep{muennighoff2022sgpt} & 1.3B & unknown & 500K & 0.320 & 0.682 & 0.162 & 0.730 & 0.830 & 0.473 & 0.545 \\
       SGPT-2.7B~\citep{muennighoff2022sgpt} & 2.7B & unknown & 500K & 0.339 & 0.701 & 0.166 & 0.752 & 0.848 & 0.489 & 0.561 \\
       \rowcolor{violet!10} \method-1B & 1B & 10M & 1.4M & 0.344 & 0.760 & 0.180 & 0.840 & 0.858 & \underline{0.531} & \underline{0.596} \\
       \rowcolor{violet!10} \method-2B & 2B & 10M & 1.4M & 0.351 & 0.760 & 0.199 & 0.863 & 0.828 & \textbf{0.543} & \textbf{0.600} \\
       \midrule 
       \multicolumn{8}{l}{\textbf{XL Size} (> 5B)} \\
       \midrule 
       SGPT-5.8B~\citep{muennighoff2022sgpt} & 5.8B & unknown & 500K & 0.362 & 0.747 & 0.199 & 0.849 & 0.863 & 0.539 & 0.604 \\
       LLaRA~\citep{li2023making} & 7B & 21M & 500K & 0.372 & 0.757 & 0.172 & 0.853 & --- & 0.539 & --- \\
       RepLLaMA~\citep{ma2023fine} & 7B & --- & 500K & 0.378 & 0.756 & 0.181 & 0.847 & --- & 0.541 & --- \\
       {LLM2Vec}$^*$~\citep{behnamghader2024llm2vec}& 7B & 1.2M & 1.5M & 0.393	&0.788	&0.225	&0.776&		0.852&	0.545&0.606 \\ 
       E5-Mistral$^*$~\citep{wang2023improving} & 7B & --- & 1.8M & 0.386 & 0.764 & 0.162 & 0.872 & 0.855 & 0.546 & 0.608 \\
       CPT-text-XL~\citep{neelakantan2022text} & 175B & unknown & unknown & 0.407 & 0.754 & --- & 0.649 & --- & --- & ---  \\
       \rowcolor{violet!10} \method-7B & 7B & 10M & 1.4M & 0.364 & 0.778 & 0.201 & 0.861 & 0.847 & \textbf{0.551} & \textbf{0.610} \\
       \bottomrule
    \end{tabular}
}
	\caption{Main experiments on biomedical text representation tasks in various scales. \textbf{Bold} and \underline{underline} indicate the best and second best results on average performance over the four retrieval tasks, and over all five tasks. $*$ denotes concurrent works (for reference only). $\dagger$ uses reranker distillation. $\ddagger$ employs hybrid retrieval. We highlight the \colorbox{green!15}{biomedical} or \colorbox{green!15}{scientific} domain-specific retrieval models. Notations are consistent across tables. ``PT'', ``FT'', and ``Sent. Sim.'' denote ``Pre-training'', ``Fine-tuning'', and ``Sentence Similarity'', respectively.\vspace{-1ex}}
	\label{tab:main_table}
\end{table*}

\noindent\textbf{Instruction Fine-tuning Dataset.}
To incorporate the model with a wide range of biomedical downstream tasks, we leverage a series of biomedical tasks with varying granularity, including both sentence-level \emph{medical natural language inference} (MedNLI)~\citep{mednli}, \emph{medical question pairs}~\citep{mccreery2020effective}, and passage-level biomedical QA tasks, including \emph{MedMCQA}~\citep{ben-abacha-etal-2019-overview}, \emph{StackExchange}~\citep{StackExchangeDataset}, and \emph{medical dialogues}~\citep{li2023chatdoctor}.
Besides, we also include several general-domain retrieval datasets, including MS MARCO~\cite{msmarco}, NQ~\citep{nq}, Fever~\citep{fever}, ELI5~\citep{fan-etal-2019-eli5}, and NLI~\citep{bowman-etal-2015-large}, to enhance the model's ability for relevance estimation. 
The instruction format and data conversion details are exhibited in Appendix~\ref{app:dataset}.

\noindent\textbf{Synthetic Data Augmentation with LLMs.}
To supplement the limited task types and relatively small sample sizes in labeled biomedical datasets, we employ a data augmentation approach to generate synthetic query and passage pairs. Two approaches are utilized for this generation process.

We leverage GPT-3.5 (\texttt{gpt-3.5-turbo-1106}) for \emph{instance-level} augmentation to enrich (query, passage) pairs resembling standard biomedical information retrieval (IR) formats. Given a passage from PubMed and Meadow used in contrastive pre-training, we prompt GPT-3.5 to generate a relevant query based on the passage context.
This allows the model to better capture the relevance within biomedical contexts for effective retrieval.

Beyond relevance signals, task generalization is also crucial for building a general retriever, as user intent and input formats vary while public data captures only a fraction of tasks. 
To address this, we perform \emph{task-level} augmentation, which involves prompting GPT-4 (\texttt{gpt-4-turbo-1106}) to conceptualize a diverse list of potential scenarios for biomedical retrieval tasks~\citep{wang2023improving}. Subsequently, we prompt GPT-4 again to generate examples for each scenario, including a query, a relevant (positive) passage, and a challenging irrelevant (hard negative) passage.
This approach allows us to enhance the diversity of instructions.

\noindent \textbf{Hard Negative Mining and Data Filter.}
In both labeled instruction fine-tuning datasets and data-label synthetic datasets, positive pairs are available, while negative examples are missing.
To obtain the negatives, we randomly select 1 passage from the top 100 passages retrieved by \texttt{E5-base}~\citep{e5} when using the given query to search the entire corpus of the corresponding dataset. 
As the generated synthetic data can be noisy, consistency filtering is adopted to filter low-quality pairs~\citep{alberti-etal-2019-synthetic}, where for each synthetic (query $q$, passage $p$) pair, we use the \texttt{E5-base} to predict the most relevant passages for ${q}$. We only retain $q$ when $p$ occurs among the top three retrieved passages.

\noindent \textbf{Fine-tuning Objectives.} 
After constructing positive and negative text pairings $\{(q_i, p_i^+, p_i^-)\}_{i=1}^{M}$ where $p_i^+$ and $p_i^-$ stands for the positive passage and the hard negative, respectively, we employ the InfoNCE loss function for each minibatch $\cB$ as 
\begin{equation}
\setlength{\abovedisplayskip}{5.2pt}
\setlength{\belowdisplayskip}{5.2pt}
\mathcal{L}_{\text{ft}}=\frac{e^{\operatorname{sim}(q_i, p_i^{+}) / \tau}}{\sum_{j \in \mathcal{B}} e^{\operatorname{sim}(q_i, p_j^{+}) / \tau}+e^{\operatorname{sim}(q_i, p_j^{-}) / \tau}},
\end{equation}
where both in-batch negatives and hard negatives are utilized to further improve model training.

\section{Experimental Results}
\label{sec:exp}

\subsection{Experimental Setups}
\noindent \textbf{Tasks and Datasets.}
We conduct experiments on eleven datasets across five biomedical retrieval-oriented tasks, including (1) IR, (2) sentence similarity (STS), (3) QA, (4) entity linking, and (5) paper recommendation.
There is \textit{no overlap} between the training and test pairs. 
Task and dataset details are available in Appendix~\ref{app:dataset}.

\noindent \textbf{Baselines.}
We compare to sparse retrieval models \emph{BM25}~\citep{bm25} and \emph{open-source} dense retrieval models with varying model sizes: 
\emph{Contriever}~\citep{izacard2022unsupervised},
\emph{Dragon}~\citep{lin-etal-2023-train},
\emph{SciMult}~\citep{zhang-etal-2023-pre},
\emph{SPECTER 2.0}~\citep{singh-etal-2023-scirepeval}, 
\emph{COCO-DR}~\citep{yu-etal-2022-coco},
\emph{QExt}~\citep{meng2022augtriever},
\emph{SGPT}~\citep{muennighoff2022sgpt},
\emph{MedCPT}~\citep{jin2023medcpt},
\emph{GTR}~\citep{ni-etal-2022-large},
\emph{InstructOR}~\citep{su-etal-2023-one},
\emph{E5-Large-v2}~\citep{e5},
\emph{BGE-Large}~\citep{chen2024bge},
\emph{LLaRA}~\citep{li2023making},
\emph{RepLLaMA}~\citep{ma2023fine},
\emph{LLM2Vec}~\citep{behnamghader2024llm2vec}, 
\emph{E5-Mistral}~\citep{wang2023improving},
and \emph{CPT-text}~\citep{neelakantan2022text}.
The details of baselines and parameter sizes are in Appendix~\ref{app:baseline}.

\noindent \textbf{Evaluation.}
We use nDCG@10 to measure standard IR performance and Spearman correlation for STS based on \emph{cosine similarity}. To evaluate the retrieval performance of QA, we report Recall@\{5,20\} and nDCG@20. For entity linking, we report mean reciprocal rank (MRR)@5 and Recall@\{1,5\}. For paper recommendation, we follow~\citet{singh-etal-2023-scirepeval} and report mean average precision (MAP) and nDCG.
The \textbf{implementation details} can be found in Appendix~\ref{app:implementation_details}.

\begin{table*}[t!]
\centering
\renewcommand\arraystretch{0.96}
\resizebox{\linewidth}{!}{
    \begin{tabular}{@{\hskip3pt}l|cc@{\hskip4pt}c|cc@{\hskip4pt}c|cc@{\hskip4pt}c|cc@{\hskip4pt}c|cc@{\hskip4pt}c|c@{\hskip5pt}c@{\hskip3pt}}
       \toprule
       \bf Task & \multicolumn{9}{c|}{\bf Question Answering} & \multicolumn{6}{c|}{\bf Entity Linking} & \multicolumn{2}{c}{\bf Paper Rec.}\\
       \midrule
         \bf \multirow{2}{*}{Model}  & \multicolumn{3}{c|}{\bf BioASQ} & \multicolumn{3}{c|}{\bf PubMedQA} & \multicolumn{3}{c|}{\bf iCliniq} & \multicolumn{3}{c|}{\bf DrugBank} & \multicolumn{3}{c|}{\bf MeSH} & \multicolumn{2}{c}{\bf RELISH} \\
          & R@5 & R@20 & nDCG@20 & R@5 & R@20 & nDCG@20 & R@5 & R@20 & nDCG@20 & R@1 & R@5 & MRR@5 & R@1 & R@5 & MRR@5 & MAP & nDCG \\
       \midrule 
       \multicolumn{10}{l}{\textbf{Base Size} (< 1B)}\\
       \midrule 
       Dragon~\shortcite{lin-etal-2023-train}  & 36.2 & \textbf{54.6} & 49.1 & \underline{71.8} & 74.0 & 72.0 & 50.6 & 65.2 & 47.4 & 81.0 & 87.6 & \underline{83.3} & 28.2 & 47.0 & 34.8 & 72.6 & 80.6 \\
        \rowcolor{green!15}MedCPT~\shortcite{jin2023medcpt}  & 34.7 & \underline{54.4} & 45.2 & 66.3 & 71.1 & 60.4 & 26.8 & 42.0 & 24.9 & 75.1 & \underline{88.0} & 80.6 & 27.7 & \underline{54.2} & 37.4 & 83.6 & 89.7\\
       E5-Large-v2$^\dagger$~\shortcite{e5} & \underline{36.8} & 54.0 & \underline{50.4} & 71.6 & \underline{74.2} & \underline{72.2} & \underline{57.6} & \underline{72.0} & \underline{55.8} & \textbf{81.8} & 86.5 & 81.5 & \textbf{32.8} & \textbf{55.0} & \textbf{41.3} & \underline{84.9} & \underline{91.0} \\
       \rowcolor{violet!10} \method-410M & \textbf{39.9} & 54.2 & \textbf{53.1} & \textbf{73.8} & \textbf{74.6} & \textbf{72.4} & \textbf{60.6} & \textbf{72.8} & \textbf{56.6} & \underline{81.4} & \textbf{88.2} & \textbf{83.7} & \underline{31.5} & 53.8 & \underline{39.8} & \textbf{85.2} & \textbf{91.2} \\
       \midrule 
       \multicolumn{8}{l}{\textbf{Large Size} (1B - 5B)}\\
      \midrule
       InstructOR-XL~\shortcite{su-etal-2023-one} & 29.9 & 43.2 & 41.8 & 70.5 & 74.0 & 69.1 & \underline{64.9} & \underline{78.1} & \underline{58.3} & 75.3 & 84.2 & 80.3 & 33.6 & 56.2 & 45.7 & 84.5 & 90.6\\
       SGPT-2.7B~\shortcite{muennighoff2022sgpt} & 33.9 & 47.4 & 47.3 & 68.3 & 73.7 & 63.2 & 45.0 & 52.2 & 41.2 & 71.9 & 77.0 & 62.9 & 20.2 & 39.7 & 28.5 & 84.9 & 90.8 \\
       \rowcolor{violet!10} \method-1B & \underline{40.4} & \underline{55.8} & \underline{53.4} & \underline{73.6} & \underline{74.4} & \underline{72.7} & 61.1 & 73.7 & 56.8 & \textbf{84.7} & \underline{89.1} & \textbf{86.5} & \underline{35.5} & \underline{60.3} & \underline{48.8} & \underline{85.2} & \underline{91.3} \\
       \rowcolor{violet!10} \method-2B & \textbf{42.5} & \textbf{56.5} & \textbf{55.7} & \textbf{74.0} & \textbf{74.6} & \textbf{73.1} & \textbf{70.0} & \textbf{81.2} & \textbf{65.7} & \underline{82.6} & \textbf{90.2} & \underline{85.8} & \textbf{45.6} & \textbf{71.3} & \textbf{59.5} & \textbf{85.4} & \textbf{91.5} \\
       \midrule 
       \multicolumn{8}{l}{\textbf{XL Size} (> 5B)} \\
       \midrule 
       E5-Mistral$^*$~\shortcite{wang2023improving} & 39.6 & 55.4 & 52.7 & 72.6 & 74.2 & 70.0 & 56.7 & 72.2 & 51.8 & 78.5 & 92.2 & 84.0 & 47.9 & 76.2 & \textbf{61.3} & 85.2 & 90.8 \\
       \rowcolor{violet!10} \method-7B & \textbf{43.7} & \textbf{60.2} & \textbf{57.4} & \textbf{74.2} & \textbf{74.6} & \textbf{73.8} & \textbf{68.4} & \textbf{79.7} & \textbf{63.7} & \textbf{84.7} & \textbf{92.8} & \textbf{88.0} & \textbf{49.8} & \textbf{76.5} & 61.1 & \textbf{86.7} & \textbf{92.2} \\
       \bottomrule
    \end{tabular}
}
	\caption{Experiments on retrieval-oriented biomedical NLP applications compared with strongest and fair baselines.\vspace{-1ex}}
	\label{tab:other_table}
\end{table*}

\subsection{Results on Text Representation Tasks}

Table~\ref{tab:main_table} presents a comprehensive evaluation of the embedding quality on four standard biomedical IR tasks and an additional task focused on biomedical sentence similarity.
Across different scales, \method outperforms the majority of baseline methods, achieving either the highest or second-highest performance in terms of average scores on the four IR tasks, as well as on the combined set of all five tasks.
It even outperforms E5-Large-v2~\citep{e5} with additional supervision signals and matches BGE-Large's hybrid retrieval approach combining dense, lexical, and multi-vector retrieval~\citep{chen2024bge}.
Here we focus on scaling up biomedical retrieval models with mixed data types, leaving the combination of \method with other more complex and larger scale language systems for future work.

A notable aspect of \method is its efficiency and lightweight nature. 
Its 410M, 1B, and 2B variants achieve 94.1\%, 97.7\%, and 98.4\% performance using only 5.9\%, 14.3\%, and 28.6\% of 7B variant's parameters, respectively. Moreover, \method-410M outperforms all the baselines in large size (1B-5B) with up to $11.7\times$ more parameters, and \method-2B matches performance with baselines in XL size (> 5B). 
Remarkably, \method also provides a reasonable training setup within an academic budget, requiring only 10M pre-training data and 1.5M fine-tuning data, which is significantly less than the data usage in most baselines, such as GTR~\citep{ni-etal-2022-large} and MedCPT~\citep{jin2023medcpt}. Yet, \method still outperforms these data-intensive methods.

\subsection{Results on Retrieval-Oriented Biomedical Applications}

Table~\ref{tab:other_table} evaluates \method's performance on biomedical downstream applications. The results demonstrate \method's efficacy over most baselines across different tasks and datasets, justifying the adaptability of our learned representations to various retrieval-oriented applications.

Furthermore, our proposed \method exhibits strong generalization capabilities across diverse tasks and input formats, including retrieving long context from short questions (BioASQ, PubMedQA), retrieving long answers from patient questions (iCliniq), retrieving definitions from entity names (DrugBank, MeSH), and retrieving relevant abstracts given an abstract (RELISH). 
Notably, \method performs well on unseen tasks, such as entity linking and paper recommendation, verifying its ability to generalize to new tasks unseen in the instruction fine-tuning stage.

\begin{figure*}[t]
    \centering    \includegraphics[width=0.93\linewidth]{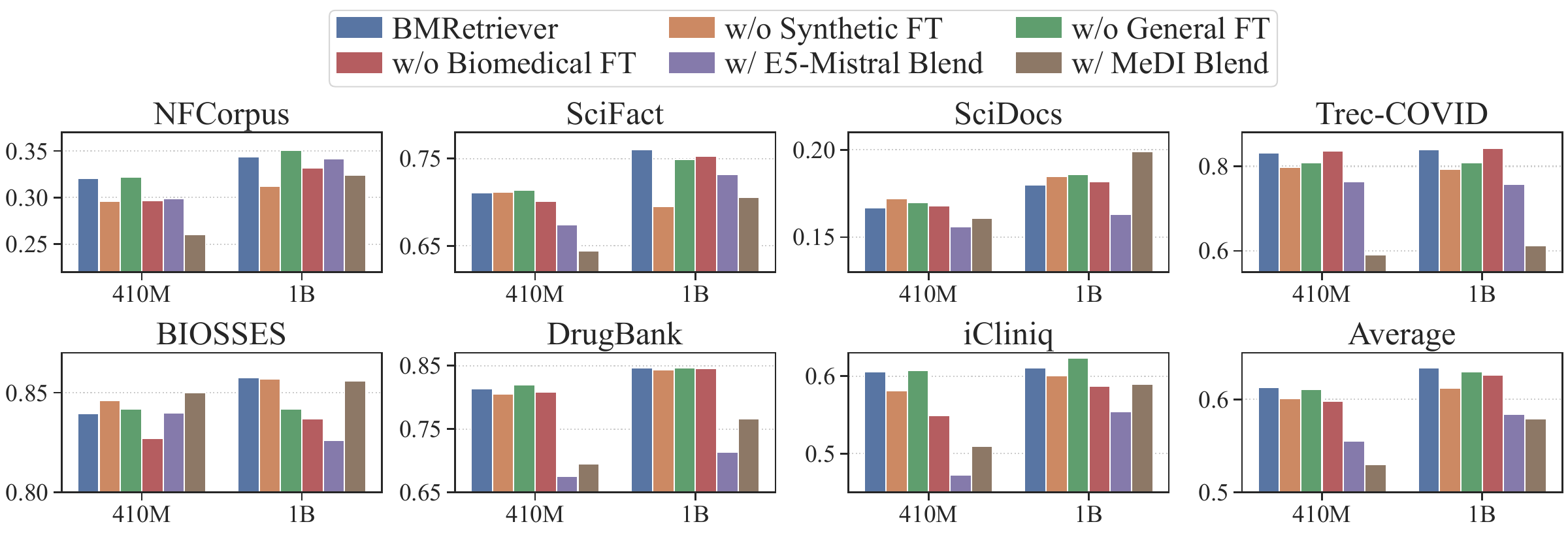}
    \vspace{-0.8ex}
    \caption{Effect of different fine-tuning data on various datasets. ``FT'' denotes ``Fine-tuning''.}
    \label{fig:diff_data_blend_per_task}
\end{figure*}

\subsection{Unsupervised Retrieval Performance}

\begin{table}[t!]
\centering
\renewcommand\arraystretch{0.96}
\resizebox{1\columnwidth}{!}{
    \begin{tabular}{l|c|ccccc|cc}
       \toprule
         \bf \multirow{2}{*}{Task} & \bf \multirow{4}{*}{Size}& \multicolumn{4}{c|}{\bf \multirow{2}{*}{Standard IR}} & \bf Sent. \\
         & & \multicolumn{4}{c|}{} & \bf Sim. & \bf Avg. & \bf Avg.\\
         \cmidrule{1-1} \cmidrule{3-7}
         \bf \multirow{2}{*}{Model} &  & \bf \multirow{2}{*}{NFC.} & \bf {Sci-} & \bf {Sci-} & \bf {Trec-} & \bf {BIO-} & \bf Retr. & \bf All \\
          & & & \bf Fact & \bf Docs & \bf COVID & \bf SSES &  &   \\
       \midrule
       Contriever~\shortcite{izacard2022unsupervised} & 110M & 0.328 & 0.677 & 0.165 & 0.274 & 0.781 &0.347 &	0.434 \\
       COCO-DR~\shortcite{yu-etal-2022-coco} & 110M & 0.243 &	0.724 &	0.150	 & 0.483 & 0.801 & 0.400 & 0.480\\
       QExt~\shortcite{meng2022augtriever} & 110M & 0.303 & 0.644 & 0.147 & 0.535 & --- & 0.407 & ---
       \\
       E5-Large-v2~\shortcite{e5} & 335M & 0.337 &	0.723	&0.218	& 0.618 & 0.822 & 0.474 & 0.543 \\
       LLM2Vec$^*$~\shortcite{behnamghader2024llm2vec} & 7B & 0.271	&0.687	&0.153&	0.557&	0.832&	0.417&	0.500 \\
       \rowcolor{violet!10} \method & 410M & 0.306 &	0.677 &	0.180 &	0.802	 &0.834	 &0.491	 & 0.560  \\
       \rowcolor{violet!10} \method & 1B &0.330 &	0.744	&0.187 &	0.800&	0.833	& 0.515	&0.579 \\
       \rowcolor{violet!10} \method & 2B & 0.342	&0.738&	0.198	&0.848&	0.847&	\underline{0.531} &	\underline{0.593} \\
       \rowcolor{violet!10} \method & 7B & 0.355 &	0.750 &	0.208  &	0.833 &	0.861 &	\textbf{0.537}	& \textbf{0.601} \\
       \bottomrule
    \end{tabular}
}
	\caption{The performance of unsupervised dense retrieval models on biomedical representation tasks. 
    Directly using the backbone model of \method{} (before contrastive pre-training) leads to performance $<0.03$ for all datasets, thus we do not report them. \vspace{-2ex}
 }
	\label{tab:unsupervised_table}
\end{table}

To highlight the effectiveness of our contrastive pre-training approach, we evaluate the performance of unsupervised dense retrieval models that only use unlabeled corpora for pre-training and synthetic data for finetuning. As shown in Table~\ref{tab:unsupervised_table}, our model outperforms existing unsupervised baselines and even surpasses many fully supervised models reported in Table~\ref{tab:main_table}.
The strong unsupervised results have important implications for real-world biomedical applications, where curating large labeled datasets is often prohibitively expensive and time-consuming. 
Our approach presents an attractive alternative, enabling the development of high-quality retrieval models in a data-efficient manner.

\begin{figure}[t]
	\centering
        \vspace{-1.5ex}
        \hspace{-0.9ex}
        \subfigure[Effect of CL strategies]{
		\includegraphics[width=0.46\linewidth]{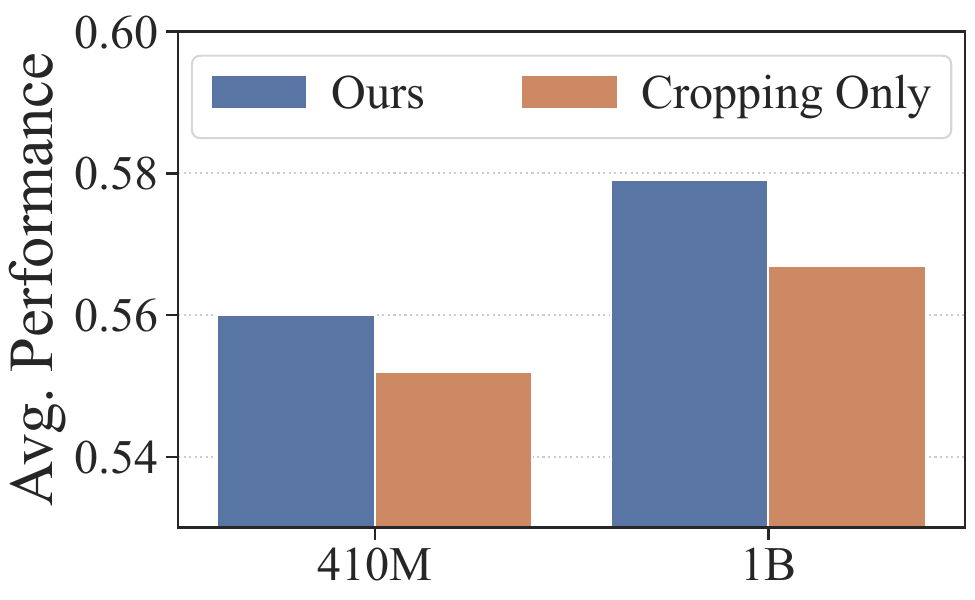}
		\label{fig:diff_cl_strategies}
	}
         \hspace{-0.9ex}
	\subfigure[Ablation Studies]{
		\includegraphics[width=0.46\linewidth]{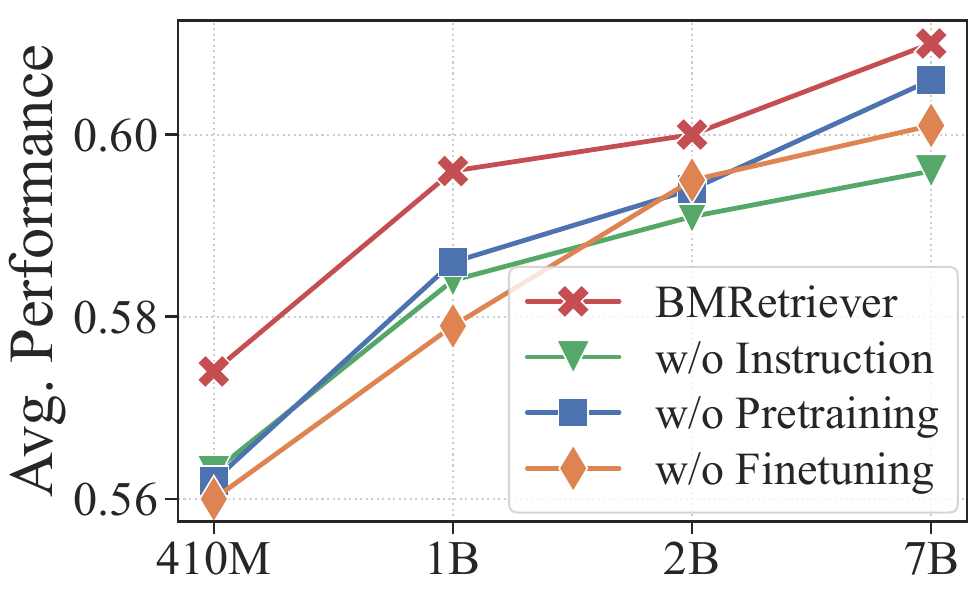}
		\label{fig:abla_ins_pre_fin}
	} 
        \hspace{-1ex}
	\caption{Additional results over five tasks in the main experiments. ``CL'' stands for ``Contrastive Learning''.\vspace{-2ex}}
\label{fig:ablation}
\end{figure}

We further investigate the performance of employing cropping alone as the contrastive pre-training strategy, which entails randomly selecting two passages from the corpus as a positive query-passage pair~\citep{gao-callan-2022-unsupervised,izacard2022unsupervised}. The results presented in Table~\ref{fig:diff_cl_strategies} demonstrate that utilizing cropping as the sole contrastive learning objective yields suboptimal performance.

\subsection{Studies on Instruction Fine-tuning}
Figure~\ref{fig:diff_data_blend_per_task} illustrates the impact of different fine-tuning data sources on model performance across various datasets\footnote{Removing biomedical data retains the synthetic data.}. Among all the utilized data types, \textit{synthetic data} contributes the most significant performance gain, which can be attributed to its larger volume compared to biomedical data and its coverage of a more diverse range of task types. It is particularly beneficial for NFCorpus, SciFact, and Trec-COVID, as these datasets follow the standard IR format of short queries and long passages, aligning with the format of the synthetic data. Furthermore, synthetic data proves advantageous for the iCliniq dataset, as it potentially includes various retrieval scenarios, such as dialog data.
\textit{General domain fine-tuning data}, consisting of short queries and long passages, generally enhances relevance estimation capabilities, benefiting standard biomedical IR tasks like Trec-COVID.
However, it may slightly diminish performance on datasets like iCliniq, which contain conversational patient-provider data that deviates from traditional IR formats, potentially introducing distribution shifts.
\textit{Biomedical fine-tuning data}, on the other hand, demonstrates broad usefulness, particularly for BIOSSES and iCliniq tasks, likely due to its inclusion of sentence similarity and dialog data aligning with downstream task formats. 
\emph{Public fine-tuning data} from \textit{E5-Mistral}~\citep{wang2023improving} and \textit{MEDI}, used by InstructOR~\citep{su-etal-2023-one}, achieves relatively poor performance, possibly due to their focus on the general domain with limited biomedical knowledge. 
Additionally, the lack of publicly available synthetic generated data from E5-Mistral~\citep{wang2023improving} may contribute to the suboptimal performance observed.

\subsection{Additional Studies}

\begin{table}[t!]
\centering
\resizebox{0.9\linewidth}{!}{
    \begin{tabular}{lcccc}
       \toprule
       \bf Stage ($\downarrow$) & \bf Volume ($\rightarrow$) & 10\% & 50\% & 100\% \\
       \midrule 
        \multirow{2}{*}{\bf Pre-training} & \method-410M & 0.540 & 0.554 & 0.560 \\
        & \method-1B & 0.564 & 0.575 & 0.579 \\
        \midrule 
        \multirow{2}{*}{\bf Fine-tuning} & \method-410M & 0.562 & 0.571 & 0.574 \\
        & \method-1B & 0.590 & 0.595 & 0.596 \\
        \bottomrule
    \end{tabular}
}
	\caption{Effect of data volume in pre-training and fine-tuning. Pre-training results do not involve subsequent fine-tuning. Fine-tuning results are based on the pre-training checkpoints with full pre-training data.\vspace{-2ex}}
	\label{tab:ft_data_volume}
\end{table}

\begin{table*}[t!]
\centering
\renewcommand\arraystretch{0.96}
\resizebox{\linewidth}{!}{
    \begin{tabular}{p{2.5cm}p{22.5cm}}
       \toprule
       \multicolumn{2}{c}{\bf Case Study 1} \\
       \midrule
       \bf Dataset / Task & iCliniq / Given a patient query, retrieve its corresponding answer from the community website.\\
       \midrule 
        \bf Query & My grandmother is having \textcolor{blue}{dizziness, vomiting and headache}. These problems occur constantly with a \textcolor{blue}{gap of five to six months}. \\
        \midrule
        \bf \method & From the symptoms that you mentioned it seems your grand mother has \textcolor{green}{migraine} which reccurs when she is either exposed to the allergen, which triggers her migraine, or it can also manifest as \textcolor{green}{a seasonal change}. How long has she has been having this problem? During her migraine episodes, how long does it last? I will recommend the following medicines :- 1) Tablet \textcolor{green}{Crocin pain relief} one tablet twice daily after food. 2) Tablet \textcolor{green}{Ondem (Ondansetron)} 4 mg, if there is severe nausea two tablets can be taken at once. Make sure her BP (blood pressure) is under control.  \\
        \midrule
        \bf \method w/o Instruction & Please provide details regarding your \textcolor{green}{headache} and \textcolor{red}{abuse}. Also, provide details about your current symptoms and like how they started and how they progress over time. What are the medications you are taking currently and for how long? Once I receive this information, I will be able to help you. \\
        \midrule
        \bf MedCPT & Since \textcolor{red}{the movement is worsening the symptoms}, it must be an inner ear problem. There could be additionally middle ear block or eustachian tube block as the pressure change worsening it. However, since you have not mentioned the duration, I am mentioning all possible differentials. Recurrent vertigo lasting under one minute is usually benign paroxysmal positional vertigo. A single episode of vertigo lasting several minutes to hours may be due to migraine or transient ischemia of the labyrinth or brainstem. $\dotsm$\\
        \midrule
        \bf E5-Mistral & You are suffering from \textcolor{green}{Episodes of headache with nausea}, which are not present every day. also \textcolor{red}{having left eye pain and twisting with tingling, imbalance while getting up, tingling in throat and over genitals, coldness over forehead ...} I would like to suggest to do some routine blood investigation tests and this symptoms seems like vitamin B12 deficiency or hypothyroidism. Are you Vegetarian?  \\
       \toprule
        \multicolumn{2}{c}{\bf Case Study 2} \\
       \midrule
       \bf Dataset / Task & SciFact / Given a biomedical claim, retrieve biomedical literature to support or refute the claim.\\
       \midrule 
        \bf Query & \textcolor{blue}{Forkhead 0 (fox0) transcription factors} are involved in \textcolor{blue}{apoptosis}. \\
        \midrule
        \bf \method & Cell Stem Cell Review \textcolor{green}{FoxO Transcription Factors} and Stem Cell Homeostasis: Insights from the Hematopoietic System \textcolor{green}{The forkhead O (FoxO) family of transcription factors} participates in diverse physiologic processes, including induction of cell-cycle arrest, stress resistance, differentiation, \textcolor{green}{apoptosis}, and metabolism. Several recent studies indicate that FoxO-dependent signaling is required for long-term regenerative potential of the hematopoietic stem cell (HSC) compartment through regulation of HSC response to physiologic oxidative stress, quiescence, and survival. $\dotsm$ \\
        \midrule
        \bf MedCPT & 
        \textcolor{green}{Forkhead box transcription factor, class O (FOXO)} is a mammalian homologue of DAF-16, which is known to regulate the lifespan of Caenorhabditis elegans and includes subfamilies of forkhead transcription factors such as AFX, FKHRL1, and FKHR. FKHR is phosphorylated on three sites (Thr-24, Ser-256, and Ser-319) in a phosphatidylinositol 3-kinase (PI3K)/Akt-dependent manner, thereby inhibiting death signals. We here documented dephosphorylation of FKHR following transient forebrain ischemia with its concomitant translocation into the nucleus in neurons in gerbil and mouse brains. The activation of FKHR preceded delayed \textcolor{red}{neuronal death in the vulnerable hippocampal regions following ischemic brain injury}. $\dotsm$ \\
        \midrule
        \makecell[lt]{\bf E5-Mistral \\ \bf \& \\ \bf \method \\ \bf w/o Instruction} & Novel Foxo1-dependent transcriptional programs control Treg cell function Regulatory T (Treg) cells, characterized by expression of the transcription factor forkhead box P3 (Foxp3), maintain immune homeostasis by suppressing self-destructive immune responses. Foxp3 operates as a late-acting differentiation factor controlling \textcolor{red}{Treg cell homeostasis and function}, whereas the early Treg-cell-lineage commitment is regulated by the Akt kinase and \textcolor{green}{the forkhead box O (Foxo) family of transcription factors}. However, whether Foxo proteins act beyond the Treg-cell-commitment stage to control Treg cell homeostasis and function remains largely unexplored. Here we show that Foxo1 is a pivotal regulator of \textcolor{red}{Treg cell function}. $\dotsm$ \\
       \bottomrule

    \end{tabular}
    
}
	\caption{A case study with two examples illustrating the quality of retrieved passages from \method compared with baseline models. \textcolor{blue}{Blue} text denotes keywords present in the original query, while \textcolor{green}{green} and \textcolor{red}{red} represent relevant and irrelevant keywords, respectively, in the retrieved passages. ``$\dotsm$'' at the end indicates that the remaining portion of the passage is omitted due to space constraints.}
 \vspace{-2.3ex}
	\label{tab:retrieve_case_study}
\end{table*}

\noindent \textbf{Ablation Studies.} We inspect different components of \method in Figure~\ref{fig:abla_ins_pre_fin}, including instruction, pre-training, and fine-tuning\footnote{Removing instruction solely eliminates the retrieval guidance, while retaining both pre-training and fine-tuning.}. The results indicate that removing any component would hurt the performance. We also observe that pre-training is particularly beneficial for smaller models, as larger models may already possess sufficient capacity to capture domain knowledge.

\noindent \textbf{Effect of Data Volume.} 
Table~\ref{tab:ft_data_volume} evaluates the effect of data volume during pre-training and fine-tuning. The results demonstrate the remarkable efficiency of \method, achieving comparable performance even when trained on substantially less data. Notably, using only 10\% of the data, the 1B variant of \method outperforms all baselines in either the pre-training or fine-tuning stage, while the 410M variant also achieves better performance than most baselines in fine-tuning.

\subsection{Case Study}

We present two case studies in Table~\ref{tab:retrieve_case_study} illustrating the quality of retrieved passages from \method compared to strong baselines. The first example, from the iCliniq dataset, considers a patient query and retrieves the corresponding answer from a community website.
In the given example, \method retrieves a passage directly addressing symptoms like \emph{headaches} and \emph{nausea}, recommending medication aligning with the condition. 
In contrast, the retrieved passage from MedCPT focuses on \emph{inner ear problems} and \emph{vertigo}, not covering the vomiting or the specific periodicity of the episodes described in the query. 
The passage from E5-Mistral talks about symptoms not mentioned by the patient, such as \emph{left eye pain} and \emph{tingling}.
Besides, we also present the result from \method without using instructions, which is also imprecise since it mentions \emph{abuse}, a topic not relevant to the query.

The second example involves retrieving biomedical literature to support or refute a claim about \emph{apoptosi}. The passage retrieved by \method specifically mentions that the \emph{FoxO} family of transcription factors participates in \emph{apoptosis}. 
Although the passage retrieved by MedCPT discusses the role of FoxO transcription factors in \emph{cell death}, it is specific to neuronal cells under \emph{ischemic conditions}, rather than general apoptosis. Furthermore, both E5-Mistral and \method without instructions retrieve an irrelevant passage about the role of FoxO1 in regulating \emph{regulatory T cells}, unrelated to the claim. 
We further illustrate the cosine similarity distributions of relevant and irrelevant (query, passage) pairs in Appendix~\ref{sec:sim}.

\vspace{-0.5ex}
\section{Conclusion}
\label{sec:conclusion}
We present \method, a series of dense retrieval models designed for knowledge-intensive biomedical NLP tasks with various scales. \method is pre-trained on a large-scale biomedical corpus and further instruction fine-tuned on diverse, high-quality biomedical tasks.
Through extensive experimentation, we have demonstrated that \method exhibits state-of-the-art performance across a range of biomedical applications. Furthermore, \method demonstrates impressive parameter efficiency, with its smaller variants achieving 94-98\% of the performance of the 7B model using only 6-29\% as many parameters, while the 410M version surpasses larger baselines (1B-5B) up to 11.7 times larger.
We hope \method can be incorporated into a broad suite of biomedical tasks to advance biomedical NLP research.

\section*{Limitation}
\noindent \textbf{Efficiency.}
One specific caveat for scaling up model size is the increment in the latency overhead. We have reported both the \emph{passage indexing} speed and \emph{retrieval latency} in Appendix \ref{app:efficiency}, which indicates that our model does not incur much additional time when compared to models with similar size (e.g., \method{}-2B  v.s. InstructOR-1.5B).
One important future work is to explore how to reduce the inference latency and lower the storage cost for text embeddings produced by LLMs.

\noindent \textbf{Cost Estimation.} Generating synthetic data using GPT models incurs additional costs. In our work, the total API cost of \method is less than \$500\footnote{As of April 2024.}, which remains affordable within an academic budget.
This cost is significantly lower than recent works~\citep{wang2023improving}, which have an estimated cost of more than \$6000. 


\section*{Ethics Consideration}

\noindent \textbf{Misinformation.} One specific issue for LLM-generated biomedical text is the potential for misinformation and hallucination~\citep{pal-etal-2023-med}.
It is important to note that for the \emph{generated queries}, the majority are short sentences or phrases without presenting any scientific facts.
We randomly selected 200 examples and asked medical students to evaluate the factuality of the generated text. The evaluation results did not reveal any misinformation or hallucination in the selected examples.

\noindent \textbf{Data Contamination.} 
A potential issue is test set contamination~\citep{sainz-etal-2023-nlp}, where some test examples overlap with the training data. This can be especially problematic for text generated by LLMs, as they are often pre-trained on massive corpora spanning various domains.
To address this concern, we follow~\citet{wang2023improving} to conduct a string match-based analysis between the test set and our training set, where we \emph{do not} observe any overlap between the train and test queries. 
While some of the corpora (e.g., PubMed) are also utilized in the test tasks, this is a standard practice even in zero-shot or few-shot evaluation of retrieval models~\citep{wang-etal-2022-gpl,yu-etal-2022-coco}, and it is not considered as contamination.

\section*{Acknowledgement}
We thank the anonymous reviewers and area chairs for valuable feedbacks. 
This research was also partially supported by the National Science Foundation under Award Number 2319449 and Award Number 2312502, the National Institute Of Diabetes And Digestive And Kidney Diseases of the National Institutes of Health under Award Number K25DK135913, the Emory Global Diabetes Center of the Woodruff Sciences Center, Emory University.
JH was supported by NSF grants IIS-1838200 and IIS-2145411. YY and CZ was supported in part by NSF IIS-2008334 and CAREER IIS-2144338.

\bibliography{anthology,custom,citation_beir}

\begin{thebibliography}{81}
\providecommand{\natexlab}[1]{#1}

\bibitem[{Alberti et~al.(2019)Alberti, Andor, Pitler, Devlin, and Collins}]{alberti-etal-2019-synthetic}
Chris Alberti, Daniel Andor, Emily Pitler, Jacob Devlin, and Michael Collins. 2019.
\newblock \href {https://doi.org/10.18653/v1/P19-1620} {Synthetic {QA} corpora generation with roundtrip consistency}.
\newblock In \emph{Proceedings of the 57th Annual Meeting of the Association for Computational Linguistics}, pages 6168--6173, Florence, Italy. Association for Computational Linguistics.

\bibitem[{Asai et~al.(2023)Asai, Schick, Lewis, Chen, Izacard, Riedel, Hajishirzi, and Yih}]{asai-etal-2023-task}
Akari Asai, Timo Schick, Patrick Lewis, Xilun Chen, Gautier Izacard, Sebastian Riedel, Hannaneh Hajishirzi, and Wen-tau Yih. 2023.
\newblock \href {https://doi.org/10.18653/v1/2023.findings-acl.225} {Task-aware retrieval with instructions}.
\newblock In \emph{Findings of the Association for Computational Linguistics: ACL 2023}, pages 3650--3675, Toronto, Canada. Association for Computational Linguistics.

\bibitem[{Bajaj et~al.(2016)Bajaj, Campos, Craswell, Deng, Gao, Liu, Majumder, McNamara, Mitra, Nguyen et~al.}]{msmarco}
Payal Bajaj, Daniel Campos, Nick Craswell, Li~Deng, Jianfeng Gao, Xiaodong Liu, Rangan Majumder, Andrew McNamara, Bhaskar Mitra, Tri Nguyen, et~al. 2016.
\newblock {MS MARCO}: A human generated machine reading comprehension dataset.
\newblock \emph{arXiv preprint arXiv:1611.09268}.

\bibitem[{BehnamGhader et~al.(2024)BehnamGhader, Adlakha, Mosbach, Bahdanau, Chapados, and Reddy}]{behnamghader2024llm2vec}
Parishad BehnamGhader, Vaibhav Adlakha, Marius Mosbach, Dzmitry Bahdanau, Nicolas Chapados, and Siva Reddy. 2024.
\newblock Llm2vec: Large language models are secretly powerful text encoders.
\newblock \emph{arXiv preprint arXiv:2404.05961}.

\bibitem[{Ben~Abacha et~al.(2019)Ben~Abacha, Shivade, and Demner-Fushman}]{ben-abacha-etal-2019-overview}
Asma Ben~Abacha, Chaitanya Shivade, and Dina Demner-Fushman. 2019.
\newblock \href {https://doi.org/10.18653/v1/W19-5039} {Overview of the {MEDIQA} 2019 shared task on textual inference, question entailment and question answering}.
\newblock In \emph{Proceedings of the 18th BioNLP Workshop and Shared Task}, pages 370--379, Florence, Italy. Association for Computational Linguistics.

\bibitem[{Biderman et~al.(2023)Biderman, Schoelkopf, Anthony, Bradley, O'Brien, Hallahan, Khan, Purohit, Prashanth, Raff, Skowron, Sutawika, and Van Der~Wal}]{pythia}
Stella Biderman, Hailey Schoelkopf, Quentin~Gregory Anthony, Herbie Bradley, Kyle O'Brien, Eric Hallahan, Mohammad~Aflah Khan, Shivanshu Purohit, Usvsn~Sai Prashanth, Edward Raff, Aviya Skowron, Lintang Sutawika, and Oskar Van Der~Wal. 2023.
\newblock \href {https://proceedings.mlr.press/v202/biderman23a.html} {Pythia: A suite for analyzing large language models across training and scaling}.
\newblock In \emph{Proceedings of the 40th International Conference on Machine Learning}, pages 2397--2430. PMLR.

\bibitem[{Boteva et~al.(2016)Boteva, Gholipour, Sokolov, and Riezler}]{nfcorpus}
Vera Boteva, Demian Gholipour, Artem Sokolov, and Stefan Riezler. 2016.
\newblock A full-text learning to rank dataset for medical information retrieval.
\newblock In \emph{European Conference on Information Retrieval}, pages 716--722. Springer.

\bibitem[{Bowman et~al.(2015)Bowman, Angeli, Potts, and Manning}]{bowman-etal-2015-large}
Samuel~R. Bowman, Gabor Angeli, Christopher Potts, and Christopher~D. Manning. 2015.
\newblock \href {https://doi.org/10.18653/v1/D15-1075} {A large annotated corpus for learning natural language inference}.
\newblock In \emph{Proceedings of the 2015 Conference on Empirical Methods in Natural Language Processing}, pages 632--642, Lisbon, Portugal. Association for Computational Linguistics.

\bibitem[{Brown et~al.(2019)Brown, Tan, El-Esawi, Liehr, Blanck, Gladue, Almeida, Cernava, Sorzano, Yeung et~al.}]{brown2019large}
Peter Brown, Aik-Choon Tan, Mohamed~A El-Esawi, Thomas Liehr, Oliver Blanck, Douglas~P Gladue, Gabriel~MF Almeida, Tomislav Cernava, Carlos~O Sorzano, Andy~WK Yeung, et~al. 2019.
\newblock Large expert-curated database for benchmarking document similarity detection in biomedical literature search.
\newblock \emph{Database}, 2019.

\bibitem[{Chen et~al.(2024)Chen, Xiao, Zhang, Luo, Lian, and Liu}]{chen2024bge}
Jianlv Chen, Shitao Xiao, Peitian Zhang, Kun Luo, Defu Lian, and Zheng Liu. 2024.
\newblock Bge m3-embedding: Multi-lingual, multi-functionality, multi-granularity text embeddings through self-knowledge distillation.
\newblock \emph{arXiv preprint arXiv:2402.03216}.

\bibitem[{Chen et~al.(2021)Chen, Allot, and Lu}]{chen2021litcovid}
Qingyu Chen, Alexis Allot, and Zhiyong Lu. 2021.
\newblock Litcovid: an open database of covid-19 literature.
\newblock \emph{Nucleic acids research}, 49(D1):D1534--D1540.

\bibitem[{Chen et~al.(2020)Chen, Ju, Dong, Fang, Wang, Yang, Zeng, Zhang, Zhang, Zhou, Zhu, and Xie}]{icliniq}
Shu Chen, Zeqian Ju, Xiangyu Dong, Hongchao Fang, Sicheng Wang, Yue Yang, Jiaqi Zeng, Ruisi Zhang, Ruoyu Zhang, Meng Zhou, Penghui Zhu, and Pengtao Xie. 2020.
\newblock \href {https://arxiv.org/abs/2004.03329} {Meddialog: {A} large-scale medical dialogue dataset}.
\newblock \emph{CoRR}, abs/2004.03329.

\bibitem[{Cohan et~al.(2020)Cohan, Feldman, Beltagy, Downey, and Weld}]{scidocs}
Arman Cohan, Sergey Feldman, Iz~Beltagy, Doug Downey, and Daniel Weld. 2020.
\newblock \href {https://aclanthology.org/2020.acl-main.207} {{SPECTER}: Document-level representation learning using citation-informed transformers}.
\newblock In \emph{Proceedings of the 58th Annual Meeting of the Association for Computational Linguistics}, pages 2270--2282, Online. Association for Computational Linguistics.

\bibitem[{Deerwester et~al.(1990)Deerwester, Dumais, Furnas, Landauer, and Harshman}]{deerwester1990indexing}
Scott Deerwester, Susan~T Dumais, George~W Furnas, Thomas~K Landauer, and Richard Harshman. 1990.
\newblock Indexing by latent semantic analysis.
\newblock \emph{Journal of the American society for information science}, 41(6):391--407.

\bibitem[{Fan et~al.(2019)Fan, Jernite, Perez, Grangier, Weston, and Auli}]{fan-etal-2019-eli5}
Angela Fan, Yacine Jernite, Ethan Perez, David Grangier, Jason Weston, and Michael Auli. 2019.
\newblock \href {https://doi.org/10.18653/v1/P19-1346} {{ELI}5: Long form question answering}.
\newblock In \emph{Proceedings of the 57th Annual Meeting of the Association for Computational Linguistics}, pages 3558--3567, Florence, Italy. Association for Computational Linguistics.

\bibitem[{Fiorini et~al.(2018)Fiorini, Leaman, Lipman, and Lu}]{fiorini2018user}
Nicolas Fiorini, Robert Leaman, David~J Lipman, and Zhiyong Lu. 2018.
\newblock How user intelligence is improving pubmed.
\newblock \emph{Nature biotechnology}, 36(10):937--945.

\bibitem[{Frisoni et~al.(2022)Frisoni, Mizutani, Moro, and Valgimigli}]{frisoni-etal-2022-bioreader}
Giacomo Frisoni, Miki Mizutani, Gianluca Moro, and Lorenzo Valgimigli. 2022.
\newblock \href {https://doi.org/10.18653/v1/2022.emnlp-main.390} {{B}io{R}eader: a retrieval-enhanced text-to-text transformer for biomedical literature}.
\newblock In \emph{Proceedings of the 2022 Conference on Empirical Methods in Natural Language Processing}, pages 5770--5793, Abu Dhabi, United Arab Emirates. Association for Computational Linguistics.

\bibitem[{Gao and Callan(2022)}]{gao-callan-2022-unsupervised}
Luyu Gao and Jamie Callan. 2022.
\newblock \href {https://doi.org/10.18653/v1/2022.acl-long.203} {Unsupervised corpus aware language model pre-training for dense passage retrieval}.
\newblock In \emph{Proceedings of the 60th Annual Meeting of the Association for Computational Linguistics (Volume 1: Long Papers)}, pages 2843--2853, Dublin, Ireland. Association for Computational Linguistics.

\bibitem[{Gillick et~al.(2019)Gillick, Kulkarni, Lansing, Presta, Baldridge, Ie, and Garcia-Olano}]{gillick-etal-2019-learning}
Daniel Gillick, Sayali Kulkarni, Larry Lansing, Alessandro Presta, Jason Baldridge, Eugene Ie, and Diego Garcia-Olano. 2019.
\newblock \href {https://doi.org/10.18653/v1/K19-1049} {Learning dense representations for entity retrieval}.
\newblock In \emph{Proceedings of the 23rd Conference on Computational Natural Language Learning (CoNLL)}, pages 528--537, Hong Kong, China. Association for Computational Linguistics.

\bibitem[{Gururangan et~al.(2020)Gururangan, Marasovi{\'c}, Swayamdipta, Lo, Beltagy, Downey, and Smith}]{gururangan-etal-2020-dont}
Suchin Gururangan, Ana Marasovi{\'c}, Swabha Swayamdipta, Kyle Lo, Iz~Beltagy, Doug Downey, and Noah~A. Smith. 2020.
\newblock \href {https://doi.org/10.18653/v1/2020.acl-main.740} {Don{'}t stop pretraining: Adapt language models to domains and tasks}.
\newblock In \emph{Proceedings of the 58th Annual Meeting of the Association for Computational Linguistics}, pages 8342--8360, Online. Association for Computational Linguistics.

\bibitem[{Hu et~al.(2022)Hu, yelong shen, Wallis, Allen-Zhu, Li, Wang, Wang, and Chen}]{hu2022lora}
Edward~J Hu, yelong shen, Phillip Wallis, Zeyuan Allen-Zhu, Yuanzhi Li, Shean Wang, Lu~Wang, and Weizhu Chen. 2022.
\newblock \href {https://openreview.net/forum?id=nZeVKeeFYf9} {Lo{RA}: Low-rank adaptation of large language models}.
\newblock In \emph{International Conference on Learning Representations}.

\bibitem[{Huang et~al.(2013)Huang, He, Gao, Deng, Acero, and Heck}]{huang2013learning}
Po{-}Sen Huang, Xiaodong He, Jianfeng Gao, Li~Deng, Alex Acero, and Larry~P. Heck. 2013.
\newblock \href {https://doi.org/10.1145/2505515.2505665} {Learning deep structured semantic models for web search using clickthrough data}.
\newblock In \emph{22nd {ACM} International Conference on Information and Knowledge Management, CIKM'13, San Francisco, CA, USA, October 27 - November 1, 2013}, pages 2333--2338. {ACM}.

\bibitem[{Izacard et~al.(2022)Izacard, Caron, Hosseini, Riedel, Bojanowski, Joulin, and Grave}]{izacard2022unsupervised}
Gautier Izacard, Mathilde Caron, Lucas Hosseini, Sebastian Riedel, Piotr Bojanowski, Armand Joulin, and Edouard Grave. 2022.
\newblock \href {https://openreview.net/forum?id=jKN1pXi7b0} {Unsupervised dense information retrieval with contrastive learning}.
\newblock \emph{Transactions on Machine Learning Research}.

\bibitem[{Jiang et~al.(2023)Jiang, Drummond, and Cohn}]{jiang-etal-2023-noisy}
Fan Jiang, Tom Drummond, and Trevor Cohn. 2023.
\newblock \href {https://doi.org/10.18653/v1/2023.findings-emnlp.803} {Noisy self-training with synthetic queries for dense retrieval}.
\newblock In \emph{Findings of the Association for Computational Linguistics: EMNLP 2023}, pages 11991--12008, Singapore. Association for Computational Linguistics.

\bibitem[{Jin et~al.(2021)Jin, Pan, Oufattole, Weng, Fang, and Szolovits}]{textbook}
Di~Jin, Eileen Pan, Nassim Oufattole, Wei-Hung Weng, Hanyi Fang, and Peter Szolovits. 2021.
\newblock What disease does this patient have? a large-scale open domain question answering dataset from medical exams.
\newblock \emph{Applied Sciences}, 11(14):6421.

\bibitem[{Jin et~al.(2019)Jin, Dhingra, Liu, Cohen, and Lu}]{jin-etal-2019-pubmedqa}
Qiao Jin, Bhuwan Dhingra, Zhengping Liu, William Cohen, and Xinghua Lu. 2019.
\newblock \href {https://doi.org/10.18653/v1/D19-1259} {{P}ub{M}ed{QA}: A dataset for biomedical research question answering}.
\newblock In \emph{Proceedings of the 2019 Conference on Empirical Methods in Natural Language Processing and the 9th International Joint Conference on Natural Language Processing (EMNLP-IJCNLP)}, pages 2567--2577, Hong Kong, China. Association for Computational Linguistics.

\bibitem[{Jin et~al.(2023)Jin, Kim, Chen, Comeau, Yeganova, Wilbur, and Lu}]{jin2023medcpt}
Qiao Jin, Won Kim, Qingyu Chen, Donald~C Comeau, Lana Yeganova, W~John Wilbur, and Zhiyong Lu. 2023.
\newblock Medcpt: Contrastive pre-trained transformers with large-scale pubmed search logs for zero-shot biomedical information retrieval.
\newblock \emph{Bioinformatics}, 39(11):btad651.

\bibitem[{Karpukhin et~al.(2020)Karpukhin, Oguz, Min, Lewis, Wu, Edunov, Chen, and Yih}]{dpr}
Vladimir Karpukhin, Barlas Oguz, Sewon Min, Patrick Lewis, Ledell Wu, Sergey Edunov, Danqi Chen, and Wen-tau Yih. 2020.
\newblock \href {https://aclanthology.org/2020.emnlp-main.550} {Dense passage retrieval for open-domain question answering}.
\newblock In \emph{Proceedings of the 2020 Conference on Empirical Methods in Natural Language Processing (EMNLP)}, pages 6769--6781, Online. Association for Computational Linguistics.

\bibitem[{Khashabi et~al.(2021)Khashabi, Ng, Khot, Sabharwal, Hajishirzi, and Callison-Burch}]{khashabi-etal-2021-gooaq-open}
Daniel Khashabi, Amos Ng, Tushar Khot, Ashish Sabharwal, Hannaneh Hajishirzi, and Chris Callison-Burch. 2021.
\newblock \href {https://doi.org/10.18653/v1/2021.findings-emnlp.38} {{G}oo{AQ}: Open question answering with diverse answer types}.
\newblock In \emph{Findings of the Association for Computational Linguistics: EMNLP 2021}, pages 421--433, Punta Cana, Dominican Republic. Association for Computational Linguistics.

\bibitem[{Kwiatkowski et~al.(2019)Kwiatkowski, Palomaki, Redfield, Collins, Parikh, Alberti, Epstein, Polosukhin, Devlin, Lee, Toutanova, Jones, Kelcey, Chang, Dai, Uszkoreit, Le, and Petrov}]{nq}
Tom Kwiatkowski, Jennimaria Palomaki, Olivia Redfield, Michael Collins, Ankur Parikh, Chris Alberti, Danielle Epstein, Illia Polosukhin, Jacob Devlin, Kenton Lee, Kristina Toutanova, Llion Jones, Matthew Kelcey, Ming-Wei Chang, Andrew~M. Dai, Jakob Uszkoreit, Quoc Le, and Slav Petrov. 2019.
\newblock \href {https://aclanthology.org/Q19-1026} {Natural questions: A benchmark for question answering research}.
\newblock \emph{Transactions of the Association for Computational Linguistics}, 7:452--466.

\bibitem[{Labrak et~al.(2024)Labrak, Bazoge, Morin, Gourraud, Rouvier, and Dufour}]{labrak2024biomistral}
Yanis Labrak, Adrien Bazoge, Emmanuel Morin, Pierre-Antoine Gourraud, Mickael Rouvier, and Richard Dufour. 2024.
\newblock Biomistral: A collection of open-source pretrained large language models for medical domains.
\newblock \emph{arXiv preprint arXiv:2402.10373}.

\bibitem[{L{\'a}la et~al.(2023)L{\'a}la, O'Donoghue, Shtedritski, Cox, Rodriques, and White}]{lala2023paperqa}
Jakub L{\'a}la, Odhran O'Donoghue, Aleksandar Shtedritski, Sam Cox, Samuel~G Rodriques, and Andrew~D White. 2023.
\newblock Paperqa: Retrieval-augmented generative agent for scientific research.
\newblock \emph{arXiv preprint arXiv:2312.07559}.

\bibitem[{Lee et~al.(2019)Lee, Chang, and Toutanova}]{lee-etal-2019-latent}
Kenton Lee, Ming-Wei Chang, and Kristina Toutanova. 2019.
\newblock \href {https://doi.org/10.18653/v1/P19-1612} {Latent retrieval for weakly supervised open domain question answering}.
\newblock In \emph{Proceedings of the 57th Annual Meeting of the Association for Computational Linguistics}, pages 6086--6096, Florence, Italy. Association for Computational Linguistics.

\bibitem[{Lewis et~al.(2020)Lewis, Perez, Piktus, Petroni, Karpukhin, Goyal, K\"{u}ttler, Lewis, Yih, Rockt\"{a}schel, Riedel, and Kiela}]{lewis2020retrieval}
Patrick Lewis, Ethan Perez, Aleksandra Piktus, Fabio Petroni, Vladimir Karpukhin, Naman Goyal, Heinrich K\"{u}ttler, Mike Lewis, Wen-tau Yih, Tim Rockt\"{a}schel, Sebastian Riedel, and Douwe Kiela. 2020.
\newblock \href {https://proceedings.neurips.cc/paper/2020/file/6b493230205f780e1bc26945df7481e5-Paper.pdf} {Retrieval-augmented generation for knowledge-intensive nlp tasks}.
\newblock In \emph{Advances in Neural Information Processing Systems}, volume~33, pages 9459--9474. Curran Associates, Inc.

\bibitem[{Li et~al.(2023{\natexlab{a}})Li, Liu, Xiao, and Shao}]{li2023making}
Chaofan Li, Zheng Liu, Shitao Xiao, and Yingxia Shao. 2023{\natexlab{a}}.
\newblock Making large language models a better foundation for dense retrieval.
\newblock \emph{arXiv preprint arXiv:2312.15503}.

\bibitem[{Li et~al.(2023{\natexlab{b}})Li, Li, Zhang, Dan, Jiang, and Zhang}]{li2023chatdoctor}
Yunxiang Li, Zihan Li, Kai Zhang, Ruilong Dan, Steve Jiang, and You Zhang. 2023{\natexlab{b}}.
\newblock Chatdoctor: A medical chat model fine-tuned on a large language model meta-ai (llama) using medical domain knowledge.
\newblock \emph{Cureus}, 15(6).

\bibitem[{Lin et~al.(2023)Lin, Asai, Li, Oguz, Lin, Mehdad, Yih, and Chen}]{lin-etal-2023-train}
Sheng-Chieh Lin, Akari Asai, Minghan Li, Barlas Oguz, Jimmy Lin, Yashar Mehdad, Wen-tau Yih, and Xilun Chen. 2023.
\newblock \href {https://doi.org/10.18653/v1/2023.findings-emnlp.423} {How to train your dragon: Diverse augmentation towards generalizable dense retrieval}.
\newblock In \emph{Findings of the Association for Computational Linguistics: EMNLP 2023}, pages 6385--6400, Singapore. Association for Computational Linguistics.

\bibitem[{Lipscomb(2000)}]{lipscomb2000medical}
Carolyn~E Lipscomb. 2000.
\newblock Medical subject headings (mesh).
\newblock \emph{Bulletin of the Medical Library Association}, 88(3):265.

\bibitem[{Liu et~al.(2021)Liu, Shareghi, Meng, Basaldella, and Collier}]{liu-etal-2021-self}
Fangyu Liu, Ehsan Shareghi, Zaiqiao Meng, Marco Basaldella, and Nigel Collier. 2021.
\newblock \href {https://doi.org/10.18653/v1/2021.naacl-main.334} {Self-alignment pretraining for biomedical entity representations}.
\newblock In \emph{Proceedings of the 2021 Conference of the North American Chapter of the Association for Computational Linguistics: Human Language Technologies}, pages 4228--4238, Online. Association for Computational Linguistics.

\bibitem[{Lo et~al.(2020)Lo, Wang, Neumann, Kinney, and Weld}]{lo-etal-2020-s2orc}
Kyle Lo, Lucy~Lu Wang, Mark Neumann, Rodney Kinney, and Daniel Weld. 2020.
\newblock \href {https://doi.org/10.18653/v1/2020.acl-main.447} {{S}2{ORC}: The semantic scholar open research corpus}.
\newblock In \emph{Proceedings of the 58th Annual Meeting of the Association for Computational Linguistics}, pages 4969--4983, Online. Association for Computational Linguistics.

\bibitem[{Loshchilov and Hutter(2019)}]{loshchilov2018decoupled}
Ilya Loshchilov and Frank Hutter. 2019.
\newblock \href {https://openreview.net/forum?id=Bkg6RiCqY7} {Decoupled weight decay regularization}.
\newblock In \emph{International Conference on Learning Representations}.

\bibitem[{Luo et~al.(2022)Luo, Mitra, Gokhale, and Baral}]{luo2022improving}
Man Luo, Arindam Mitra, Tejas Gokhale, and Chitta Baral. 2022.
\newblock Improving biomedical information retrieval with neural retrievers.
\newblock In \emph{Proceedings of the AAAI Conference on Artificial Intelligence}, volume~36, pages 11038--11046.

\bibitem[{Ma et~al.(2023)Ma, Wang, Yang, Wei, and Lin}]{ma2023fine}
Xueguang Ma, Liang Wang, Nan Yang, Furu Wei, and Jimmy Lin. 2023.
\newblock Fine-tuning llama for multi-stage text retrieval.
\newblock \emph{arXiv preprint arXiv:2310.08319}.

\bibitem[{McCreery et~al.(2020)McCreery, Katariya, Kannan, Chablani, and Amatriain}]{mccreery2020effective}
Clara~H McCreery, Namit Katariya, Anitha Kannan, Manish Chablani, and Xavier Amatriain. 2020.
\newblock Effective transfer learning for identifying similar questions: matching user questions to covid-19 faqs.
\newblock In \emph{Proceedings of the 26th ACM SIGKDD international conference on knowledge discovery \& data mining}, pages 3458--3465.

\bibitem[{Meng et~al.(2022)Meng, Liu, Yavuz, Agarwal, Tu, Yu, Zhang, Bhat, and Zhou}]{meng2022augtriever}
Rui Meng, Ye~Liu, Semih Yavuz, Divyansh Agarwal, Lifu Tu, Ning Yu, Jianguo Zhang, Meghana Bhat, and Yingbo Zhou. 2022.
\newblock Augtriever: Unsupervised dense retrieval by scalable data augmentation.
\newblock \emph{arXiv preprint arXiv:2212.08841}.

\bibitem[{Mohan et~al.(2017)Mohan, Fiorini, Kim, and Lu}]{mohan-etal-2017-deep}
Sunil Mohan, Nicolas Fiorini, Sun Kim, and Zhiyong Lu. 2017.
\newblock \href {https://doi.org/10.18653/v1/W17-2328} {Deep learning for biomedical information retrieval: Learning textual relevance from click logs}.
\newblock In \emph{{B}io{NLP} 2017}, pages 222--231, Vancouver, Canada,. Association for Computational Linguistics.

\bibitem[{Muennighoff(2022)}]{muennighoff2022sgpt}
Niklas Muennighoff. 2022.
\newblock Sgpt: Gpt sentence embeddings for semantic search.
\newblock \emph{arXiv preprint arXiv:2202.08904}.

\bibitem[{Naik et~al.(2022)Naik, Parasa, Feldman, Wang, and Hope}]{naik-etal-2022-literature}
Aakanksha Naik, Sravanthi Parasa, Sergey Feldman, Lucy~Lu Wang, and Tom Hope. 2022.
\newblock \href {https://doi.org/10.18653/v1/2022.findings-naacl.33} {Literature-augmented clinical outcome prediction}.
\newblock In \emph{Findings of the Association for Computational Linguistics: NAACL 2022}, pages 438--453, Seattle, United States. Association for Computational Linguistics.

\bibitem[{Neelakantan et~al.(2022)Neelakantan, Xu, Puri, Radford, Han, Tworek, Yuan, Tezak, Kim, Hallacy et~al.}]{neelakantan2022text}
Arvind Neelakantan, Tao Xu, Raul Puri, Alec Radford, Jesse~Michael Han, Jerry Tworek, Qiming Yuan, Nikolas Tezak, Jong~Wook Kim, Chris Hallacy, et~al. 2022.
\newblock Text and code embeddings by contrastive pre-training.
\newblock \emph{arXiv preprint arXiv:2201.10005}.

\bibitem[{Ni et~al.(2022)Ni, Qu, Lu, Dai, Hernandez~Abrego, Ma, Zhao, Luan, Hall, Chang, and Yang}]{ni-etal-2022-large}
Jianmo Ni, Chen Qu, Jing Lu, Zhuyun Dai, Gustavo Hernandez~Abrego, Ji~Ma, Vincent Zhao, Yi~Luan, Keith Hall, Ming-Wei Chang, and Yinfei Yang. 2022.
\newblock \href {https://doi.org/10.18653/v1/2022.emnlp-main.669} {Large dual encoders are generalizable retrievers}.
\newblock In \emph{Proceedings of the 2022 Conference on Empirical Methods in Natural Language Processing}, pages 9844--9855, Abu Dhabi, United Arab Emirates. Association for Computational Linguistics.

\bibitem[{Pal et~al.(2023)Pal, Umapathi, and Sankarasubbu}]{pal-etal-2023-med}
Ankit Pal, Logesh~Kumar Umapathi, and Malaikannan Sankarasubbu. 2023.
\newblock \href {https://doi.org/10.18653/v1/2023.conll-1.21} {{M}ed-{HALT}: Medical domain hallucination test for large language models}.
\newblock In \emph{Proceedings of the 27th Conference on Computational Natural Language Learning (CoNLL)}, pages 314--334, Singapore. Association for Computational Linguistics.

\bibitem[{Raffel et~al.(2020)Raffel, Shazeer, Roberts, Lee, Narang, Matena, Zhou, Li, and Liu}]{raffel2020exploring}
Colin Raffel, Noam Shazeer, Adam Roberts, Katherine Lee, Sharan Narang, Michael Matena, Yanqi Zhou, Wei Li, and Peter~J Liu. 2020.
\newblock Exploring the limits of transfer learning with a unified text-to-text transformer.
\newblock \emph{Journal of machine learning research}, 21(140):1--67.

\bibitem[{Rasley et~al.(2020)Rasley, Rajbhandari, Ruwase, and He}]{rasley2020deepspeed}
Jeff Rasley, Samyam Rajbhandari, Olatunji Ruwase, and Yuxiong He. 2020.
\newblock Deepspeed: System optimizations enable training deep learning models with over 100 billion parameters.
\newblock In \emph{Proceedings of the 26th ACM SIGKDD International Conference on Knowledge Discovery \& Data Mining}, pages 3505--3506.

\bibitem[{Robertson et~al.(2009)Robertson, Zaragoza et~al.}]{bm25}
Stephen Robertson, Hugo Zaragoza, et~al. 2009.
\newblock The probabilistic relevance framework: Bm25 and beyond.
\newblock \emph{Foundations and Trends{\textregistered} in Information Retrieval}, 3(4):333--389.

\bibitem[{Sainz et~al.(2023)Sainz, Campos, Garc{\'\i}a-Ferrero, Etxaniz, de~Lacalle, and Agirre}]{sainz-etal-2023-nlp}
Oscar Sainz, Jon Campos, Iker Garc{\'\i}a-Ferrero, Julen Etxaniz, Oier~Lopez de~Lacalle, and Eneko Agirre. 2023.
\newblock \href {https://doi.org/10.18653/v1/2023.findings-emnlp.722} {{NLP} evaluation in trouble: On the need to measure {LLM} data contamination for each benchmark}.
\newblock In \emph{Findings of the Association for Computational Linguistics: EMNLP 2023}, pages 10776--10787, Singapore. Association for Computational Linguistics.

\bibitem[{Shi et~al.(2023)Shi, Zhuang, Zhu, Iwinski, Wattenbarger, and Wang}]{shi2023retrieval}
Wenqi Shi, Yuchen Zhuang, Yuanda Zhu, Henry Iwinski, Michael Wattenbarger, and May~Dongmei Wang. 2023.
\newblock Retrieval-augmented large language models for adolescent idiopathic scoliosis patients in shared decision-making.
\newblock In \emph{Proceedings of the 14th ACM International Conference on Bioinformatics, Computational Biology, and Health Informatics}, pages 1--10.

\bibitem[{Shivade(2017)}]{mednli}
Chaitanya Shivade. 2017.
\newblock \href {https://doi.org/10.13026/C2RS98} {Mednli — a natural language inference dataset for the clinical domain}.

\bibitem[{Singh et~al.(2023)Singh, D{'}Arcy, Cohan, Downey, and Feldman}]{singh-etal-2023-scirepeval}
Amanpreet Singh, Mike D{'}Arcy, Arman Cohan, Doug Downey, and Sergey Feldman. 2023.
\newblock \href {https://doi.org/10.18653/v1/2023.emnlp-main.338} {{S}ci{R}ep{E}val: A multi-format benchmark for scientific document representations}.
\newblock In \emph{Proceedings of the 2023 Conference on Empirical Methods in Natural Language Processing}, pages 5548--5566, Singapore. Association for Computational Linguistics.

\bibitem[{So{\u{g}}anc{\i}o{\u{g}}lu et~al.(2017)So{\u{g}}anc{\i}o{\u{g}}lu, {\"O}zt{\"u}rk, and {\"O}zg{\"u}r}]{souganciouglu2017biosses}
Gizem So{\u{g}}anc{\i}o{\u{g}}lu, Hakime {\"O}zt{\"u}rk, and Arzucan {\"O}zg{\"u}r. 2017.
\newblock Biosses: a semantic sentence similarity estimation system for the biomedical domain.
\newblock \emph{Bioinformatics}, 33(14):i49--i58.

\bibitem[{Su et~al.(2023)Su, Shi, Kasai, Wang, Hu, Ostendorf, Yih, Smith, Zettlemoyer, and Yu}]{su-etal-2023-one}
Hongjin Su, Weijia Shi, Jungo Kasai, Yizhong Wang, Yushi Hu, Mari Ostendorf, Wen-tau Yih, Noah~A. Smith, Luke Zettlemoyer, and Tao Yu. 2023.
\newblock \href {https://doi.org/10.18653/v1/2023.findings-acl.71} {One embedder, any task: Instruction-finetuned text embeddings}.
\newblock In \emph{Findings of the Association for Computational Linguistics: ACL 2023}, pages 1102--1121, Toronto, Canada. Association for Computational Linguistics.

\bibitem[{Team(2021)}]{StackExchangeDataset}
Flax Sentence~Embeddings Team. 2021.
\newblock \href {https://huggingface.co/datasets/flax-sentence-embeddings/} {Stack exchange question pairs}.

\bibitem[{Team et~al.(2024)Team, Mesnard, Hardin, Dadashi, Bhupatiraju, Pathak, Sifre, Rivi{\`e}re, Kale, Love et~al.}]{team2024gemma}
Gemma Team, Thomas Mesnard, Cassidy Hardin, Robert Dadashi, Surya Bhupatiraju, Shreya Pathak, Laurent Sifre, Morgane Rivi{\`e}re, Mihir~Sanjay Kale, Juliette Love, et~al. 2024.
\newblock Gemma: Open models based on gemini research and technology.
\newblock \emph{arXiv preprint arXiv:2403.08295}.

\bibitem[{Thakur et~al.(2021)Thakur, Reimers, R{\"u}ckl{\'e}, Srivastava, and Gurevych}]{beir}
Nandan Thakur, Nils Reimers, Andreas R{\"u}ckl{\'e}, Abhishek Srivastava, and Iryna Gurevych. 2021.
\newblock \href {https://openreview.net/forum?id=wCu6T5xFjeJ} {{BEIR}: A heterogeneous benchmark for zero-shot evaluation of information retrieval models}.
\newblock In \emph{Thirty-fifth Conference on Neural Information Processing Systems Datasets and Benchmarks Track (Round 2)}.

\bibitem[{Thorne et~al.(2018)Thorne, Vlachos, Christodoulopoulos, and Mittal}]{fever}
James Thorne, Andreas Vlachos, Christos Christodoulopoulos, and Arpit Mittal. 2018.
\newblock \href {https://aclanthology.org/N18-1074} {{FEVER}: a large-scale dataset for fact extraction and {VER}ification}.
\newblock In \emph{Proceedings of the 2018 Conference of the North {A}merican Chapter of the Association for Computational Linguistics: Human Language Technologies, Volume 1 (Long Papers)}, pages 809--819, New Orleans, Louisiana. Association for Computational Linguistics.

\bibitem[{Tsatsaronis et~al.(2015)Tsatsaronis, Balikas, Malakasiotis, Partalas, Zschunke, Alvers, Weissenborn, Krithara, Petridis, Polychronopoulos et~al.}]{bioasq}
George Tsatsaronis, Georgios Balikas, Prodromos Malakasiotis, Ioannis Partalas, Matthias Zschunke, Michael~R Alvers, Dirk Weissenborn, Anastasia Krithara, Sergios Petridis, Dimitris Polychronopoulos, et~al. 2015.
\newblock An overview of the {BIOASQ} large-scale biomedical semantic indexing and question answering competition.
\newblock \emph{BMC bioinformatics}, 16(1):1--28.

\bibitem[{Voorhees et~al.(2021)Voorhees, Alam, Bedrick, Demner-Fushman, Hersh, Lo, Roberts, Soboroff, and Wang}]{treccovid}
Ellen Voorhees, Tasmeer Alam, Steven Bedrick, Dina Demner-Fushman, William~R. Hersh, Kyle Lo, Kirk Roberts, Ian Soboroff, and Lucy~Lu Wang. 2021.
\newblock \href {https://doi.org/10.1145/3451964.3451965} {{TREC-COVID}: Constructing a pandemic information retrieval test collection}.
\newblock \emph{SIGIR Forum}, 54(1).

\bibitem[{Wadden et~al.(2020)Wadden, Lin, Lo, Wang, van Zuylen, Cohan, and Hajishirzi}]{scifact}
David Wadden, Shanchuan Lin, Kyle Lo, Lucy~Lu Wang, Madeleine van Zuylen, Arman Cohan, and Hannaneh Hajishirzi. 2020.
\newblock \href {https://aclanthology.org/2020.emnlp-main.609} {Fact or fiction: Verifying scientific claims}.
\newblock In \emph{Proceedings of the 2020 Conference on Empirical Methods in Natural Language Processing (EMNLP)}, pages 7534--7550, Online. Association for Computational Linguistics.

\bibitem[{Wang et~al.(2022{\natexlab{a}})Wang, Thakur, Reimers, and Gurevych}]{wang-etal-2022-gpl}
Kexin Wang, Nandan Thakur, Nils Reimers, and Iryna Gurevych. 2022{\natexlab{a}}.
\newblock \href {https://doi.org/10.18653/v1/2022.naacl-main.168} {{GPL}: Generative pseudo labeling for unsupervised domain adaptation of dense retrieval}.
\newblock In \emph{Proceedings of the 2022 Conference of the North American Chapter of the Association for Computational Linguistics: Human Language Technologies}, pages 2345--2360, Seattle, United States. Association for Computational Linguistics.

\bibitem[{Wang et~al.(2022{\natexlab{b}})Wang, Yang, Huang, Jiao, Yang, Jiang, Majumder, and Wei}]{e5}
Liang Wang, Nan Yang, Xiaolong Huang, Binxing Jiao, Linjun Yang, Daxin Jiang, Rangan Majumder, and Furu Wei. 2022{\natexlab{b}}.
\newblock Text embeddings by weakly-supervised contrastive pre-training.
\newblock \emph{arXiv preprint arXiv:2212.03533}.

\bibitem[{Wang et~al.(2024)Wang, Yang, Huang, Yang, Majumder, and Wei}]{wang2023improving}
Liang Wang, Nan Yang, Xiaolong Huang, Linjun Yang, Rangan Majumder, and Furu Wei. 2024.
\newblock Improving text embeddings with large language models.
\newblock \emph{arXiv preprint arXiv:2401.00368}.

\bibitem[{Wang et~al.(2020)Wang, Lo, Chandrasekhar, Reas, Yang, Burdick, Eide, Funk, Katsis, Kinney, Li, Liu, Merrill, Mooney, Murdick, Rishi, Sheehan, Shen, Stilson, Wade, Wang, Wang, Wilhelm, Xie, Raymond, Weld, Etzioni, and Kohlmeier}]{wang-etal-2020-cord}
Lucy~Lu Wang, Kyle Lo, Yoganand Chandrasekhar, Russell Reas, Jiangjiang Yang, Doug Burdick, Darrin Eide, Kathryn Funk, Yannis Katsis, Rodney~Michael Kinney, Yunyao Li, Ziyang Liu, William Merrill, Paul Mooney, Dewey~A. Murdick, Devvret Rishi, Jerry Sheehan, Zhihong Shen, Brandon Stilson, Alex~D. Wade, Kuansan Wang, Nancy Xin~Ru Wang, Christopher Wilhelm, Boya Xie, Douglas~M. Raymond, Daniel~S. Weld, Oren Etzioni, and Sebastian Kohlmeier. 2020.
\newblock \href {https://aclanthology.org/2020.nlpcovid19-acl.1} {{CORD-19}: The {COVID-19} open research dataset}.
\newblock In \emph{Proceedings of the 1st Workshop on {NLP} for {COVID-19} at {ACL} 2020}, Online. Association for Computational Linguistics.

\bibitem[{Wishart et~al.(2018)Wishart, Feunang, Guo, Lo, Marcu, Grant, Sajed, Johnson, Li, Sayeeda et~al.}]{wishart2018drugbank}
David~S Wishart, Yannick~D Feunang, An~C Guo, Elvis~J Lo, Ana Marcu, Jason~R Grant, Tanvir Sajed, Daniel Johnson, Carin Li, Zinat Sayeeda, et~al. 2018.
\newblock Drugbank 5.0: a major update to the drugbank database for 2018.
\newblock \emph{Nucleic acids research}, 46(D1):D1074--D1082.

\bibitem[{Xiong et~al.(2024)Xiong, Jin, Lu, and Zhang}]{xiong2024benchmarking}
Guangzhi Xiong, Qiao Jin, Zhiyong Lu, and Aidong Zhang. 2024.
\newblock Benchmarking retrieval-augmented generation for medicine.
\newblock \emph{arXiv preprint arXiv:2402.13178}.

\bibitem[{Xiong et~al.(2021)Xiong, Xiong, Li, Tang, Liu, Bennett, Ahmed, and Overwijk}]{ance}
Lee Xiong, Chenyan Xiong, Ye~Li, Kwok-Fung Tang, Jialin Liu, Paul~N. Bennett, Junaid Ahmed, and Arnold Overwijk. 2021.
\newblock \href {https://openreview.net/forum?id=zeFrfgyZln} {Approximate nearest neighbor negative contrastive learning for dense text retrieval}.
\newblock In \emph{International Conference on Learning Representations}.

\bibitem[{Xu et~al.(2024)Xu, Shi, Yu, Zhuang, Jin, Wang, Ho, and Yang}]{xu2024ram}
Ran Xu, Wenqi Shi, Yue Yu, Yuchen Zhuang, Bowen Jin, May~D Wang, Joyce~C Ho, and Carl Yang. 2024.
\newblock Ram-ehr: Retrieval augmentation meets clinical predictions on electronic health records.
\newblock \emph{arXiv preprint arXiv:2403.00815}.

\bibitem[{Yang et~al.(2018)Yang, Qi, Zhang, Bengio, Cohen, Salakhutdinov, and Manning}]{hotpotqa}
Zhilin Yang, Peng Qi, Saizheng Zhang, Yoshua Bengio, William Cohen, Ruslan Salakhutdinov, and Christopher~D. Manning. 2018.
\newblock \href {https://aclanthology.org/D18-1259} {{H}otpot{QA}: A dataset for diverse, explainable multi-hop question answering}.
\newblock In \emph{Proceedings of the 2018 Conference on Empirical Methods in Natural Language Processing}, pages 2369--2380, Brussels, Belgium. Association for Computational Linguistics.

\bibitem[{Yu et~al.(2024)Yu, Ping, Liu, Wang, You, Zhang, Shoeybi, and Catanzaro}]{yu2024rankrag}
Yue Yu, Wei Ping, Zihan Liu, Boxin Wang, Jiaxuan You, Chao Zhang, Mohammad Shoeybi, and Bryan Catanzaro. 2024.
\newblock Rankrag: Unifying context ranking with retrieval-augmented generation in llms.
\newblock \emph{arXiv preprint arXiv:2407.02485}.

\bibitem[{Yu et~al.(2022)Yu, Xiong, Sun, Zhang, and Overwijk}]{yu-etal-2022-coco}
Yue Yu, Chenyan Xiong, Si~Sun, Chao Zhang, and Arnold Overwijk. 2022.
\newblock \href {https://doi.org/10.18653/v1/2022.emnlp-main.95} {{COCO}-{DR}: Combating distribution shift in zero-shot dense retrieval with contrastive and distributionally robust learning}.
\newblock In \emph{Proceedings of the 2022 Conference on Empirical Methods in Natural Language Processing}, pages 1462--1479, Abu Dhabi, United Arab Emirates. Association for Computational Linguistics.

\bibitem[{Zhang et~al.(2024{\natexlab{a}})Zhang, Yu, Wang, and Zhang}]{zhang2024arl2}
Lingxi Zhang, Yue Yu, Kuan Wang, and Chao Zhang. 2024{\natexlab{a}}.
\newblock Arl2: Aligning retrievers for black-box large language models via self-guided adaptive relevance labeling.
\newblock \emph{arXiv preprint arXiv:2402.13542}.

\bibitem[{Zhang et~al.(2024{\natexlab{b}})Zhang, Patil, Jain, Shen, Zaharia, Stoica, and Gonzalez}]{zhang2024raft}
Tianjun Zhang, Shishir~G Patil, Naman Jain, Sheng Shen, Matei Zaharia, Ion Stoica, and Joseph~E. Gonzalez. 2024{\natexlab{b}}.
\newblock \href {https://openreview.net/forum?id=rzQGHXNReU} {{RAFT}: Adapting language model to domain specific {RAG}}.
\newblock In \emph{First Conference on Language Modeling}.

\bibitem[{Zhang et~al.(2023)Zhang, Cheng, Shen, Liu, Wang, and Gao}]{zhang-etal-2023-pre}
Yu~Zhang, Hao Cheng, Zhihong Shen, Xiaodong Liu, Ye-Yi Wang, and Jianfeng Gao. 2023.
\newblock \href {https://doi.org/10.18653/v1/2023.findings-emnlp.820} {Pre-training multi-task contrastive learning models for scientific literature understanding}.
\newblock In \emph{Findings of the Association for Computational Linguistics: EMNLP 2023}, pages 12259--12275, Singapore. Association for Computational Linguistics.

\end{thebibliography}

\clearpage

\appendix

\section{Additional Synthetic Data Augmentation Details}

\subsection{Prompt Format to Generate Task and Pairs}

\begin{lstlisting}[style=mystyle, caption={Prompt format for synthetic retrieval task generation.}, label=lst:prompt, escapeinside={<@}{@>}]
Brainstorm a list of potentially useful biomedical text retrieval tasks.

Here are a few examples for your reference:
1. Provided a scientific claim as query, retrieve documents that help verify or refute the claim.
2. Search for documents that answers a FAQ-style query on children's nutrition.

Please adhere to the following guidelines:
1. Specify what the query is, and what the desired documents are.
2. Each retrieval task should cover a wide range of queries, and should not be too specific.
3. Focus on biomedical related topics.

Your output should always be a python list of strings only, with about 20 elements, and each element corresponds to a distinct retrieval task in one sentence. Do not explain yourself or output anything else. Be creative!
\end{lstlisting}

\begin{lstlisting}[style=mystyle, caption={Prompt format for synthetic retrieval examples generation.}, label=lst:prompt, escapeinside={<@}{@>}]
You have been assigned a biomedical retrieval task: <@\textcolor{blue}{[task]}@>

Your mission is to write one biomedical text retrieval example for this task in JSON format. The JSON object must contain the following keys:
1. "user_query": a string, a random user search query specified by the retrieval task.
2. "positive_document": a string, a relevant document for the user query.
3. "hard_negative_document": a string, a hard negative document that only appears relevant to the query.

Please adhere to the following guidelines:
1. The "user_query" should be <@\textcolor{blue}{[query\_type]}@>, <@\textcolor{blue}{[query\_length]}@>, <@\textcolor{blue}{[clarity]}@>, and diverse in topic.
2. All documents should be at least <@\textcolor{blue}{[num\_words]}@> words long.
3. Both the query and documents should be in English.
4. Both the query and documents require <@\textcolor{blue}{[difficulty]}@> level education to understand.

Your output must always be a JSON object only, do not explain yourself or output anything else. Be creative!
\end{lstlisting}

\textcolor{blue}{\texttt{[task]}}: The task names generated from the previous step.

\textcolor{blue}{\texttt{[query\_type]}}: Randomly sampled from ["extremely long-tail", "long-tail", "common"].

\textcolor{blue}{\texttt{[query\_length]}}: Randomly sampled from ["less than 5 words", "5-10 words", "at least 10 words"]

\textcolor{blue}{\texttt{[clarity]}}: Randomly sampled from ["clear", "understandable with some effort", "ambiguous"]

\textcolor{blue}{\texttt{[num\_words]}}: Randomly sampled from ["50 words", "50-100 words", "200 words", "300 words", "400 words"]

\textcolor{blue}{\texttt{[difficulty]}}: Randomly sampled from ["high school", "college", "PhD"]

\subsection{Case Study}
\begin{table*}[t!]
\centering
\renewcommand\arraystretch{0.92}
\resizebox{\linewidth}{!}{
    \begin{tabular}{lp{16.5cm}}
       \toprule
       \multicolumn{2}{c}{\bf Synthetic Data Augmentation Example 1} \\
       \midrule
       \bf Synthetic Retrieval Task & Retrieve documents that discuss the potential health risks associated with obesity\\
       \midrule 
        \bf Query & Potential health hazards related to obesity \\
        \midrule
        \bf Positive Passage & The escalating obesity epidemic is associated with a range of serious health risks and complications. Obesity is a chronic disease characterized by excessive accumulation and storage of fat in the body, which in turn is associated with a multitude of adverse health effects. These effects include an increased risk of premature death, cardiovascular diseases including hypertension, stroke and coronary artery disease, type 2 diabetes, certain types of cancer, and sleep apnea among others. The psychosocial consequences of obesity are also significant, with reduced quality of life and depression being common. The pathophysiological basis of these complications is complex and involves a myriad of interrelated metabolic, inflammatory, and vascular processes. For instance, the excess adiposity in obesity contributes to systemic inflammation and insulin resistance, which are central to the development of type 2 diabetes and cardiovascular diseases. The increased mechanical load of obesity can also lead to degenerative joint diseases, while the increased metabolic demand can lead to heart failure. Furthermore, the excess adipose tissue can secrete a variety of adipokines and cytokines that can promote carcinogenesis. Thus, the potential health risks associated with obesity are multifold, and the prevention and management of obesity should be a public health priority. \\
        \midrule
        \bf Negative Passage & Obesity is increasingly recognized as a major health problem worldwide. While it is associated with a multitude of health risks, the potential benefits of obesity are less well known. Some studies have suggested that obese individuals may have a lower risk of certain conditions such as osteoporosis, due to their higher bone mineral density. Furthermore, some obese individuals seem to be 'metabolically healthy', with no apparent signs of the metabolic abnormalities typically associated with obesity. These findings, however, do not negate the overall detrimental effects of obesity on health. Rather, they highlight the complexity of obesity as a disease and the need for personalized approaches in its management. It is also important to note that the 'obesity paradox', where obesity seems to confer survival advantages in certain conditions such as heart failure, remains controversial and is not yet fully understood. Overall, while there may be potential benefits associated with obesity, these are likely outweighed by the substantial health risks, and efforts should be focused on preventing and managing obesity to improve health outcomes. \\
        \toprule
        \multicolumn{2}{c}{\bf Synthetic Data Augmentation Example 2} \\
        \midrule
        \bf Synthetic Retrieval Task & Search for documents that provide information on the latest treatments for autoimmune diseases\\
        \midrule 
        \bf Query & I am looking for scholarly articles or scientific papers that describe the most recent advancements in therapies for autoimmune diseases, such as rheumatoid arthritis, lupus, celiac disease, or multiple sclerosis. \\
        \midrule
        \bf Positive Passage & In recent years, there have been significant advancements in the treatment of autoimmune diseases. One major development is the use of biologics, which are drugs derived from living organisms. Biologics have been successfully used in the treatment of rheumatoid arthritis, lupus, and other autoimmune disorders. They work by targeting specific components of the immune system that cause inflammation and damage. Another promising treatment is stem cell therapy, which has potential in treating diseases such as multiple sclerosis. In this procedure, the patient's immune system is suppressed and then re-established with the patient's own stem cells, essentially 'resetting' the immune system. Moreover, dietary intervention, such as a strict gluten-free diet, has been proven to manage celiac disease effectively. However, these treatments all have their own risks and side effects, and research is ongoing to refine these therapies and develop new ones. \\
        \midrule
        \bf Negative Passage & Autoimmune disorders are a group of diseases where the body's immune system attacks its own cells. There are many types of autoimmune diseases, including Rheumatoid Arthritis, Lupus, Celiac Disease, and Multiple Sclerosis. Each of these diseases has different symptoms, causes, and requires different treatments. Some common symptoms of autoimmune diseases are fatigue, joint pain, and swelling, skin problems, and abdominal pain. The causes of these diseases are not fully understood, but they are thought to be a combination of genetic and environmental factors. There is currently no cure for autoimmune diseases, but treatments can help manage the symptoms. Treatments include medication, physical therapy, and in some cases surgery. In the case of celiac disease, a strict gluten-free diet is necessary. It is important to work with a healthcare provider to develop a treatment plan that is tailored to the individual's needs. \\
       \bottomrule
    \end{tabular}
}
	\caption{Synthetic retrieval tasks and examples generated by GPT-4.}
	\label{tab:syn_case_study}
\end{table*}

We present a list of generated retrieval scenarios as examples:

\begin{itemize}
    \item ``Search for articles discussing the latest advancements in neurology.''
    \item ``Retrieval of articles discussing the symptoms and treatments of rare diseases given a query on rare diseases.''
    \item ``Find documents that discuss the impact of lifestyle changes on a specific medical condition.''
    \item ``Locate documents that provide information on the epidemiology of a certain disease in a specific region.''
    \item \quad $\dotsi$
\end{itemize}

Table~\ref{tab:syn_case_study} presents two illustrative examples where GPT-4 generates corresponding queries, positive passages, and negative passages for each synthetic retrieval task. The complete set of task names is provided in the supplementary materials.

\section{Task and Dataset Information}
\label{app:dataset}

\subsection{Pre-training Corpus}
\label{app:pt-data}

\begin{table}[t!]
\centering
\resizebox{\linewidth}{!}{
    \begin{tabular}{lcp{5cm}}
       \toprule
       \bf Dataset & \bf Size & \bf Line \\
       \midrule 
        PubMed~\shortcite{xiong2024benchmarking} & 8M$^*$& \url{https://huggingface.co/datasets/MedRAG/pubmed} \\
        \midrule
        arXiv, MedRxiv, BioRxiv & 577K & \url{https://huggingface.co/datasets/mteb/raw_arxiv} \\
        \midrule
        Meadow~\shortcite{wang-etal-2020-cord} & 460k & \url{https://huggingface.co/datasets/medalpaca/medical_meadow_cord19} \\
        \midrule
        Textbooks~\shortcite{textbook} & 50K & \url{https://huggingface.co/datasets/MedRAG/textbooks} \\
        \midrule
        StatPearls~\shortcite{xiong2024benchmarking} & 54K & \url{https://huggingface.co/datasets/MedRAG/statpearls} \\
        \midrule
        LitCovid~\shortcite{chen2021litcovid} & 70K & \url{https://huggingface.co/datasets/KushT/LitCovid_BioCreative} \\
        \midrule
        S2ORC~\shortcite{lo-etal-2020-s2orc} & 600K & \url{https://github.com/allenai/s2orc} \\
        \midrule
        MS Marco~\shortcite{msmarco} &1.2M & \url{https://huggingface.co/datasets/Tevatron/msmarco-passage-corpus} \\
       \bottomrule
    \end{tabular}
}
	\caption{Biomedical corpora collection for unsupervised contrastive pre-training. $^*$: We randomly select 8M corpus from the full collections.
	}
	\label{tab:pt_data}
\end{table}

We publicly release the training recipe used in both the pre-training and fine-tuning stages to ensure transparency, reproducibility, and potential applicability to new domains.
To equip \method with a strong foundation in biomedical contexts, we compile a diverse corpus of biomedical data sources. 
Table~\ref{tab:pt_data} summarizes the unlabeled corpora used for contrastive pre-training of our model, including their sizes and public availability.
For pre-training on \method-7b, we only use 1M passages due to the efficiency issue.

For queries and passages, the instruction used in the contrastive pre-training stage is ``\texttt{Given a query, retrieve passages that are relevant to the query. Query: \{\}}'', ``\texttt{Represent this passage. Passage: \{\}}''.

\subsection{Fine-tuning Task and Dataset}
\label{sec:convert}
\begin{table*}[t!]
\centering
\resizebox{\linewidth}{!}{
    \begin{tabular}{p{3cm}ccp{6cm}p{6cm}}
       \toprule
       \bf Dataset & \bf Size & \bf Task & \bf Link & \bf Instruction Format \\
       \midrule 
       \multicolumn{5}{l}{\bf BioMedical Domain} \\
       \midrule
        StackExchange \shortcite{StackExchangeDataset} & 43K & QA & \url{https://huggingface.co/datasets/flax-sentence-embeddings/stackexchange_titlebody_best_voted_answer_jsonl} & Given a biological query from the stackexchange, retrieve replies most relevant to the query \\
        \midrule
        MedNLI \shortcite{mednli} & 4.6K & Sentence Similarity & \url{https://physionet.org/content/mednli/1.0.0/} & Given a sentence, retrieve sentences with the same meaning \\
        \midrule
        MQP \shortcite{mccreery2020effective} & 3K & Sentence Similarity & \url{https://huggingface.co/datasets/medical_questions_pairs} & Given a sentence, retrieve sentences with the same meaning \\
        \midrule
        MedQuad \shortcite{ben-abacha-etal-2019-overview} & 47K & QA & \url{https://huggingface.co/datasets/lavita/MedQuAD} & Given a question, retrieve relevant documents that answer the question \\
        \midrule
        HealthcareMagic \shortcite{li2023chatdoctor} & 30K & Dialogue & \url{https://huggingface.co/datasets/medical_dialog} & Given a question with context from online medical forums, retrieve responses that best answer the question \\
        \midrule
        \multicolumn{5}{l}{\bf General Domain} \\
        \midrule
        ELI5 \shortcite{fan-etal-2019-eli5} & 18K$^*$  & Longform QA & \url{https://huggingface.co/datasets/eli5} & Given a question, retrieve the highest voted answers on Reddit forum \\
        \midrule
        GooAQ \shortcite{khashabi-etal-2021-gooaq-open} & 100K$^*$  & QA & \url{https://huggingface.co/datasets/gooaq} & Given a question, retrieve relevant passages that answer the question \\
        \midrule
        MS Marco \shortcite{msmarco} & 500K& Web Search & \url{https://huggingface.co/datasets/ms_marco} & Given a web search query, retrieve relevant passages that answer the query \\
        \midrule
        NQ \shortcite{nq} & 58K & QA & \url{https://github.com/facebookresearch/DPR/blob/main/dpr/data/download_data.py} & Given a question, retrieve Wikipedia passages that answer the question \\
        \midrule
        FEVER \shortcite{fever} & 10K$^*$  & Fact Verification & \url{https://huggingface.co/datasets/BeIR/fever} & Given a claim, retrieve documents that support or refute the claim \\
        \midrule
        HotpotQA \shortcite{hotpotqa} & 8K$^*$  & Question Answering & \url{https://huggingface.co/datasets/BeIR/hotpotqa} & Given a complex question, retrieve documents that answer the question \\
        \midrule
        NLI \shortcite{bowman-etal-2015-large} & 138K$^*$  & Natural Language Inference & \url{https://github.com/princeton-nlp/SimCSE/blob/main/data/download_nli.sh} & Given a premise, retrieve hypotheses that are entailed by the premise \\
       \bottomrule
    \end{tabular}
}
	\caption{Labeled data collection for instruction fine-tuning with a diverse range of tasks, including both sentence-level NLI and passage-level QA. $^*$: Only a subset of the original dataset is sampled.
	}
	\label{tab:ft_data}
\end{table*}

\paragraph{Real Datasets.} Table~\ref{tab:ft_data} displays the datasets used for instruction fine-tuning besides synthetic augmentation, which include a diverse range of tasks at both the sentence and passage levels across biomedical and general domains. 
Biomedical datasets cover biomedical QA (\citealt{StackExchangeDataset}, \citealt{ben-abacha-etal-2019-overview}), sentence similarity (\citealt{mednli}, \citealt{mccreery2020effective}), and dialogue (\citealt{li2023chatdoctor}). General domain datasets tackle long-form QA (\citealt{fan-etal-2019-eli5}), web search~\citep{msmarco}, open-domain QA (\citealt{khashabi-etal-2021-gooaq-open}, \citealt{nq}), fact verification (\citealt{fever}), NLI (\citealt{bowman-etal-2015-large}), and web search (\citealt{msmarco}).
For MS Marco\footnote{\url{https://msmarco.z22.web.core.windows.net/msmarcoranking/triples.train.small.tar.gz}} and NQ dataset\footnote{\url{https://dl.fbaipublicfiles.com/dpr/data/retriever/biencoder-nq-adv-hn-train.json.gz}}, we use the ground-truth annotations as well as the provided hard negative to form the fine-tuning data.  

For non-retrieval tasks, we convert them into a retrieval format as follows:
\begin{itemize}
    \item For standard QA datasets, we directly use the question as the query and the gold evidence passages as the ground-truth passages.
    \item For NLI and sentence similarity tasks, we treat sentence pairs identified as "entail" or "similar" as positive examples, while those labeled as "contradict" or "non-similar" serve as \emph{hard negatives}.
    \item For medical dialogue datasets, we consider the answer to the user query as the ground-truth passage for retrieval.
\end{itemize}

The query instructions are listed in the corresponding tables, while for passages, we use the same instruction format as the template used in the pre-training stage: ``\texttt{Represent this passage. Passage: \{\}}''.

\paragraph{Synthetic Datasets.} We leverage LLM-generated synthetic data to augment the training set. For the query generation scenario, we generate 500K synthetic queries. After round-trip filtering, we retain approximately 420K (query, passage) pairs.
The instructions used for generating synthetic queries are:
\begin{itemize}
    \item For the PubMed corpus: ``\texttt{Given a question, retrieve Pubmed passages that answer the question.}''
    \item For the Meadow corpus on COVID-19: ``\texttt{Given a query on COVID-19, retrieve COVID-19 related articles that answer the query.}''
\end{itemize}

We generate 20,000 synthetic tasks and query-passage pairs using GPT-4. Table~\ref{tab:syn_case_study} presents some examples of synthetic retrieval tasks and query-passage pairs.

\subsection{Evaluation Task and Dataset}
\label{app:eva-data}

\begin{table*}[t!]
\centering
\resizebox{\linewidth}{!}{
    \begin{tabular}{lp{2.5cm}ccp{5cm}p{5cm}}
       \toprule
       \bf Dataset & \bf Task & \bf \# Queries & \bf \# Documents & \bf Link & \bf Instruction Format\\
       \midrule 
        NFCorpus \shortcite{nfcorpus} & Biomedical Search & 323 & 3.6K & \url{https://huggingface.co/datasets/BeIR/nfcorpus} & Given a question, retrieve relevant documents that best answer the question\\
        \midrule
        SciFact \shortcite{scifact}& Fact Verification & 300 & 5K & \url{https://huggingface.co/datasets/BeIR/scifact} & Given a scientific claim, retrieve documents that support or refute the claim \\
        \midrule
        SciDocs \shortcite{scidocs} & Citation Prediction & 1,000 & 25K & \url{https://huggingface.co/datasets/BeIR/scidocs} & Given a scientific paper title, retrieve paper abstracts that are cited by the given paper\\
        \midrule
        Trec-COVID \shortcite{treccovid} & Biomedical Search & 50 & 171K & \url{https://huggingface.co/datasets/BeIR/trec-covid} & Given a query on COVID-19, retrieve documents that answer the query \\
        \midrule
        BIOSSES \shortcite{souganciouglu2017biosses} & Biomedical Sentence Similarity & 100 & --- & \url{ https://huggingface.co/datasets/biosses} & Given a sentence, retrieve sentences with the same meaning\\
        \midrule
        BioASQ \shortcite{bioasq} & Biomedical QA & 500 & 500K & \url{http://participants-area.bioasq.org/datasets/} & Given a question, retrieve Pubmed passages that answer the question\\
        \midrule
        PubMedQA \shortcite{jin-etal-2019-pubmedqa} & Biomedical QA & 500 & 211K& \url{https://huggingface.co/datasets/qiaojin/PubMedQA}& Given a question, retrieve Pubmed passages that answer the question\\
        \midrule
        iCliniq \shortcite{icliniq} & Biomedical CQA & 7.3K & 7.3K & \url{https://huggingface.co/datasets/medical_dialog} & Given a question with context from online medical forums, retrieve responses that best answer the question\\
        \midrule
        DrugBank \shortcite{wishart2018drugbank} & Biomedical Entity Linking & 4.1K & 4.1K & \url{https://go.drugbank.com/}& Given a drug, retrieve passages for its definition\\
        \midrule
        MeSH \shortcite{lipscomb2000medical} & Biomedical Entity Linking & 29.6K & 29.6K & \url{https://www.nlm.nih.gov/databases/download/mesh.html} & Given a concept, retrieve passages for its definition\\
        \midrule
        RELISH \shortcite{singh-etal-2023-scirepeval,brown2019large} & Biomedical Paper Recommendation & 3.2K & {191.2K} & \url{https://huggingface.co/datasets/allenai/scirepeval/viewer/relish} & Given an article, retrieve Pubmed articles that are relevant to this article\\
       \bottomrule
    \end{tabular}
}
	\caption{Evaluation datasets for biomedical text representation tasks and retrieval-oriented downstream applications. 
	}
	\label{tab:eval_data}
\end{table*}

We conduct a comprehensive evaluation of \method on eleven datasets (Table~\ref{tab:eval_data}) across five biomedical tasks, including:

\paragraph{Information Retrieval.} For passage retrieval tasks in biomedicine, we select four datasets from the BEIR benchmark~\citep{beir}, each focusing on biomedical or scientific-related IR tasks involving complex, terminology-rich documents:
(1) \textbf{NFCorpus}~\citep{nfcorpus} contains 323 queries related to nutrition facts for medical IR, sourced from 3.6K PubMed documents;
(2) \textbf{SciFact}~\citep{scifact} includes 300 queries, aiming to retrieve evidence-containing abstracts from 5K scientific papers for fact-checking;
(3) \textbf{SciDocs}~\citep{scidocs} consists of 25K scientific papers for citation prediction with 1K queries containing article titles; 
(4) \textbf{TREC-COVID}~\citep{treccovid} includes 50 queries, with an average of 493.5 relevant documents per query, specifically curated for biomedical IR related to COVID-19.

\paragraph{Sentence Similarity.} 
For sentence retrieval tasks, we evaluate retrieval models on
(5) \textbf{BIOSSES}~\citep{souganciouglu2017biosses}, which comprises 100 sentence pairs extracted from PubMed articles. The similarity of each sentence pair is annotated using a 5-point scale, ranging from 0 (no relation) to 4 (equivalent).

\paragraph{Question-and-Answering.} Besides passage and sentence retrieval tasks, we further evaluate the effectiveness of retrieval models on several retrieval-oriented downstream tasks, including biomedical QA.
(6) \textbf{BioASQ}~\citep{bioasq} and (7) \textbf{PubMedQA}~\citep{jin-etal-2019-pubmedqa} are large-scale biomedical multi-choice QA datasets derived from PubMed articles.
(8) \textbf{iCliniq}~\citep{icliniq} contains medical QA pairs from the public health forum derived from conversations between clinicians and patients.

\paragraph{Entity Linking.}
For additional retrieval-oriented downstream applications, we conduct two biomedical entity-linking experiments:
(9) \textbf{DrugBank}~\citep{wishart2018drugbank} for drug entity matching, and (10) \textbf{MeSH}~\citep{lipscomb2000medical} for biomedical concept linking.

\paragraph{Paper Recommendation.} We evaluate the performance of retrieval models on a paper recommendation task using the (11) \textbf{RELISH} dataset~\citep{singh-etal-2023-scirepeval,brown2019large}. It assigns similarity scores ranging from 0 (not similar) to 2 (similar) for locating relevant literature from more than 180K PubMed abstracts.

\section{Baseline Information}
\label{app:baseline}
We consider both sparse and dense retrieval models to provide a comprehensive evaluation of retrieval models in biomedical applications.

\subsection{Baselines for Retrieval Tasks in Main Experiments}
\paragraph{Sparse Retrieval Models.} Sparse retrieval models rely on lexical matching between query and document terms to calculate similarity scores.

\begin{itemize}[leftmargin=0.3cm]
\item \textbf{BM25}~\citep{bm25} is the most commonly used sparse retrieval model for lexical retrieval, employing a scoring function that calculates the similarity between two high-dimensional sparse vectors based on token matching and weighting. 
\end{itemize}

\paragraph{Dense Retrieval Models.}
Dense retrieval models utilize dense vector representations to capture semantic similarity between queries and documents.
In our experiments, we consider dense retrieval models at various scales for a comprehensive evaluation: (1) \textbf{Base Size} (<1B parameters), (2) \textbf{Large Size} (1B-5B), and (3) \textbf{XL Size} (>5B).

\begin{itemize}[leftmargin=0.3cm]
\item \textbf{Contriever}~\citep{izacard2022unsupervised} is a dense retrieval model (110M) pre-trained via contrastive learning on documents sampled from Wikipedia and CC-Net  corpora.

\item \textbf{Dragon}~\citep{lin-etal-2023-train} is a BERT-base-sized dense retrieval model (110M) that undergoes progressive training using a data augmentation approach, incorporating diverse queries and sources of supervision.

\item \textbf{SPECTER 2.0}~\citep{singh-etal-2023-scirepeval} is a scientific document representation model (110M) pretrained using multi-format representation learning, enabling tailored embeddings for a diverse range of task formats.

\item \textbf{SciMult}~\citep{zhang-etal-2023-pre} is a scientific dense retrieval model (110M) that employs a multi-task contrastive learning framework with task-aware specialization and instruction tuning to enhance performance on scientific literature retrieval tasks.

\item \textbf{COCO-DR}~\citep{yu-etal-2022-coco} is a lightweight dense retrieval model (110M) pre-trained using continuous contrastive learning and implicit distributionally robust optimization on domain-specific corpora, enabling adaptation to various downstream tasks, including those in the biomedical domain.

\item \textbf{QExt}~\citep{meng2022augtriever} is a data augmentation method that trains dense retrieval models by selecting salient spans from the original document, and generating pseudo queries using transferred language models. We report the performance of QExt (110M) in the unsupervised dense retrieval setting only.

\item \textbf{SGPT}~\citep{muennighoff2022sgpt} is a dense retrieval model that employs position-weighted mean pooling and fine-tunes only bias tensors to learn effective representations for semantic search. We conduct a comprehensive comparison with SGPT at different scales, including base size (125M), large size (1.3B and 2.7B), and XL size (5.8B).

\item \textbf{MedCPT}~\citep{jin2023medcpt} is a biomedical embedding model (220M) specifically designed for biomedical literature retrieval, leveraging contrastive pre-training on medical corpora consisting of 255M user clicks from PubMed search logs~\citep{fiorini2018user}.

\item \textbf{GTR}~\citep{ni-etal-2022-large} is a generalizable dense retriever that initializes its dual encoders from T5~\citep{raffel2020exploring}. 
GTR is pre-trained on a private Community QA dataset and fine-tuned using the NQ~\citep{nq} and MS MARCO~\citep{msmarco} datasets.
We conduct a comprehensive comparison with GTR at varying scales, including GTR-Large (335M), GTR-XL (1.2B), and GTR-XXL (4.8B).

\item \textbf{InstructOR}~\citep{su-etal-2023-one} is a multitask embedder that generates task- and domain-aware embeddings for a given text input and its corresponding task instructions, without requiring any additional training.
We evaluate InstructOR at both base (335M) and large (1.5B) scales.

\item \textbf{E5-Large-v2}~\citep{e5} adopts a complex multi-stage training paradigm that first pre-trains on large-scale weakly-supervised text pairs and then fine-tunes on several labeled datasets. Due to the knowledge distillation from a teacher model, which involves additional supervised training signals, we consider E5-Large-v2 (335M) as an \emph{unfair comparison}.

\item \textbf{BGE-Large}~\citep{chen2024bge} is a dense retrieval model (335M) that uses graph-based embedding techniques and a multi-stage training paradigm similar to E5~\citep{e5}. However, since BGE employs a hybrid retrieval strategy incorporating additional lexical and multi-vector retrieval, we consider it an \emph{unfair comparison} for our focus on dense retrieval only. We acknowledge BGE as a concurrent work and provide its performance for reference purposes.

\item \textbf{LLaRA}~\citep{li2023making} uses LLM-generated text embeddings to reconstruct input sentence tokens and predict next sentence tokens.

\item \textbf{RepLLaMA}~\citep{ma2023fine} is a dense retriever (7B) that fine-tunes the LLaMA model for effective representation learning in passage and document retrieval using the MS MARCO datasets~\citep{msmarco}.

\item \textbf{LLM2Vec}~\citep{behnamghader2024llm2vec} is an unsupervised approach that transforms LLMs into text encoders by enabling bidirectional attention via masked next token prediction and adopts unsupervised contrastive learning for sequence representation learning. 

\item \textbf{E5-Mistral}~\citep{wang2023improving} is an enhanced version of the E5~\citep{e5} that incorporates synthetic data generated by LLMs for a diverse range of text embedding tasks. 

\item \textbf{CPT-text}~\citep{neelakantan2022text} is a dense retrieval model pre-trained on web-scale data using a contrastive objective with neighboring text passages as positive pairs. We only consider its performance as a reference rather than a fair comparison due to its large size (175B parameters).
\end{itemize}

\subsection{Baselines for Retrieval-Oriented Downstream Applications}
In experiments for retrieval-oriented downstream applications, we only compare \method to the strongest, most relevant, and fair baselines, including:
(1) \textbf{Base Size} (<1B): {Dragon}~\citep{lin-etal-2023-train}, {MedCPT}~\citep{jin2023medcpt}, and {E5-Large-v2}~\citep{e5};
(2) \textbf{Large Size} (1B-5B): {InstructOR}~\citep{su-etal-2023-one} and {SGPT-2.7B}~\citep{muennighoff2022sgpt}; and
(3) \textbf{XL Size} (>5B): {E5-Mistral}~\citep{wang2023improving}.

\section{Implementation Details}
\label{app:implementation_details}
The backbones used for \method are available in Table~\ref{tab:method_overview}. The learning rates are set to $5e-5$ for the 410M and 1B variants, $4e-5$ for the 2B variant, and $2e-5$ for the 7B variant during pre-training; $5e-5$ for the 410M and 1B  variants, $2e-5$ for the 2B variant, and $1e-5$ for the 7B variant during fine-tuning. The global batch size is set to $256$ for the 410M and 1B variants, $128$ for the 2B variant, and $64$ for 7B variants.
To optimize GPU memory consumption, we train our models with LoRA ($r=16$, $\alpha=32$)~\citep{hu2022lora}, brain floating point (bfloat16) quantization, and DeepSpeed gradient checkpointing~\citep{rasley2020deepspeed}. The training is performed on 4 NVIDIA H100 GPUs for 2 epochs during pre-training and 1 epoch during fine-tuning, using a maximum sequence length of 512 tokens. We use the AdamW optimizer~\citep{loshchilov2018decoupled} with a linear learning rate warm-up for the first 100 steps.
For contrastive learning, we set $\tau=1$ without any further tuning.

\section{Cosine Similarity v.s. Dot Product}
\label{sec:cos_dot}
\begin{figure}[h]
    \centering    
    \includegraphics[width=0.55\linewidth]{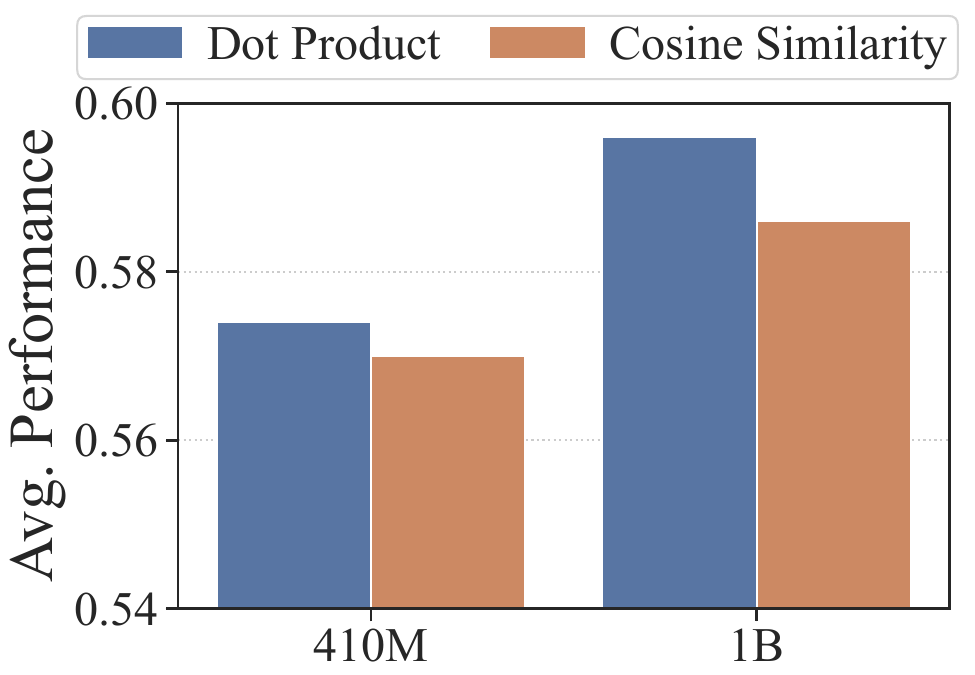}
    \caption{Comparison of performance using dot product and cosine similarity.}
    \label{fig:diff_similarity}
\end{figure}

We explore different objectives for embedding similarity, namely dot product and cosine similarity. From the experimental results in Figure~\ref{fig:diff_similarity}, we empirically observe that the dot product could achieve a better empirical performance. Thus, we choose to use dot product by default as our similarity metrics.

\section{Similarity Score}
\label{sec:sim}

\begin{figure}[h]
	\centering
        \hspace{-1ex}
        \subfigure[SciFact]{
		\includegraphics[width=0.48\linewidth]{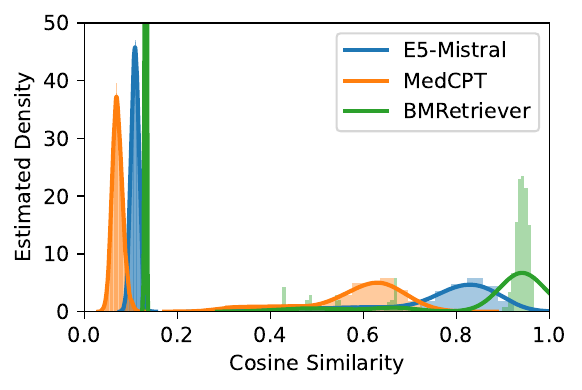}
		\label{fig:scifact_similarity}
	}
         \hspace{-3ex}
	\subfigure[iCliniq]{
		\includegraphics[width=0.48\linewidth]{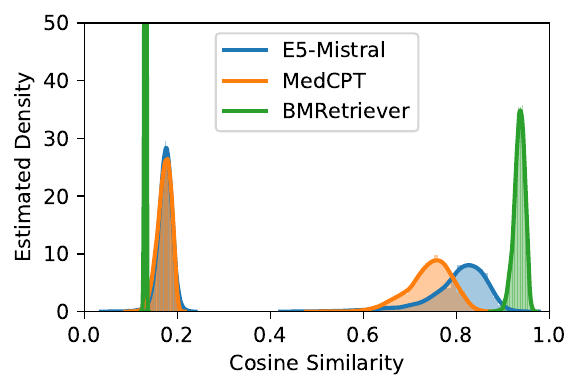}
		\label{fig:icliniq_similarity}
	} 
        \hspace{-1ex}
	\caption{The cosine similarity on positive pair embeddings and negative pair embeddings.\vspace{-2ex}}
\label{fig:simlarity}
\end{figure}

Figure~\ref{fig:simlarity} depicts the distributions of cosine similarity scores for positive and negative embedding pairs across two datasets. The left side displays the similarity distributions for negative examples, while the right side shows the distributions for positive examples. These figures illustrate that \method exhibits a larger separation between positive and negative examples, showing its enhanced ability to effectively retrieve relevant passages.

\section{Efficiency}
\label{app:efficiency}

Table \ref{tab:time_overview} exhibits the document encoding speed and retrieval latency of \method and baseline dense retrieval models. While \method introduces additional encoding latency compared to BERT-based retrievers, we do not incorporate significant overhead when compared to baselines of similar model size. 

\begin{table}[h]
\centering
\resizebox{0.91\linewidth}{!}{
    \begin{tabular}{lccc}
       \toprule
       \bf Models & \textbf{Size} & \makecell[c]{\textbf{Document Encoding Speed} \\ (\# docs / s / GPU)} & \makecell[c]{\textbf{Retrieval Latency} \\ (ms)} \\
       \midrule 
        \rowcolor{green!15}MedCPT~\shortcite{jin2023medcpt} & 220M & 1390.1 & 11.6 \\
        InstructOR~\shortcite{su-etal-2023-one} &1.5B & 181.2 & 14.6 \\
        SGPT~\shortcite{muennighoff2022sgpt} &2.7B & 98.5 & 35.5 \\
        E5-Mistral$^*$~\shortcite{wang2023improving} & 7B & 51.8 & 58.6 \\
        \rowcolor{violet!10}\method &410M & 471.2 & 14.6 \\
        \rowcolor{violet!10}\method &1B & 194.0 & 28.6 \\
        \rowcolor{violet!10}\method &2B & 166.2 & 28.6 \\
        \rowcolor{violet!10}\method &7B & 51.8 & 58.6 \\
       \bottomrule
    \end{tabular}
}
	\caption{Time complexity of \method. \vspace{-1.5ex}}
	\label{tab:time_overview}
\end{table}

\end{document}